\documentclass[Afour,sageh,times]{sagej}
\usepackage[colorlinks,bookmarksopen,bookmarksnumbered,citecolor=red,urlcolor=red]{hyperref}
\usepackage{amsmath,amsthm,amssymb,amsfonts}

\usepackage{algorithm}
\usepackage{algpseudocode}
\usepackage{array}
\usepackage[caption=false,font=normalsize,labelfont=sf,textfont=sf]{subfig}
\usepackage{textcomp}
\usepackage{stfloats}
\usepackage{url}
\usepackage{verbatim}
\usepackage{cite}
\usepackage{bm}
\usepackage{multirow}
\usepackage{textcomp}
\usepackage{siunitx}
\usepackage{arydshln}

\usepackage{wrapfig}
\usepackage{siunitx}

\usepackage{hyperref}
\usepackage{arydshln}  
\usepackage{algpseudocode}
\usepackage{amsthm}
\usepackage{cancel}

\usepackage{graphicx}      
\usepackage{caption}       
\usepackage{subcaption}    

\usepackage{epstopdf}

\usepackage{multirow}
\newcommand\BibTeX{{\rmfamily B\kern-.05em \textsc{i\kern-.025em b}\kern-.08em
T\kern-.1667em\lower.7ex\hbox{E}\kern-.125emX}}

\def\volumeyear{2016}

\begin{document}

\runninghead{Smith and Wittkopf}


\title{A Task-Driven, Planner-in-the-Loop Computational Design Framework for Modular Manipulators %
}

\author{Maolin Lei\affilnum{1}, Edoardo Romiti\affilnum{1}, Arturo Laurenzi\affilnum{1}, Rui Dai\affilnum{1}, Matteo Dalle Vedove\affilnum{1,2}, Jiatao Ding\affilnum{2},  Daniele Fontanelli\affilnum{2}, Nikos Tsagarakis\affilnum{1} }

\affiliation{\affilnum{1}Humanoids and Human Centered Mechatronics Research Line, Italian Institute of Technology (IIT) Genoa, Italy\\
\affilnum{2}Department of Industrial Engineering, University of Trento, Trento, Italy}

 \corrauth{Maolin Lei: maolin.lei@outlook.com}

\begin{abstract}
Modular manipulators, composed of pre-manufactured and interchangeable base modules, provide high adaptability across diverse task. However, deploying such systems requires generating feasible motions while simultaneously optimizing the manipulator’s morphology and mounted pose under kinematic, dynamic, and physical constraints for the given task scenarios. Moreover, traditional single-branch morphological designs often rely on increasing link length to extend reach, which is prone to exceed the torque limit of the joint module near the base link. To overcome the above challenges, we propose a unified task-driven computational framework that consists of trajectory planning across varying morphologies with the co-optimization of morphology and mounted pose. Within this framework, a hierarchical model predictive control (HMPC) strategy is developed to enable motion planning for both redundant and non-redundant manipulators under multi-subtask scenarios. For design optimization, the covariance matrix adaptation evolution strategy (CMA-ES) is employed to efficiently explore a hybrid search space comprising discrete morphology configurations and continuous mounted poses. Additionally, we introduce a virtual module abstraction to support the generation of bi-branch morphologies, allowing the auxiliary branch to offload torque from the primary branch and extend the system's capability with the larger workspace task. Extensive simulations and physical experiments across polishing, drilling, and pick-and-place tasks validate the framework’s effectiveness. Extensive simulations and hardware experiments have demonstrated the following: 1) Given a desired task such as pick-and-place, polishing or drilling this framework can generate various designs that satisfy both kinematic and dynamic constraints while avoids the environment collision; 2) By customizing objective functions, this framework allows for flexible design targeting various goals, including maximizing manipulability, minimizing joint effort, and reducing the number of modules; 3) Using this framework, we successfully designed a bi-branch morphology capable of operating in a large workspace without necessitating the manufacture of a more powerful basic module. To the best of our knowledge, this is the first work on the automatic selection between single-branch and bi-branch morphologies.

\end{abstract}

\keywords{modular manipulator, computational design, morphology and mounted pose optimization, planner in the loop optimization}

\maketitle

\section{Introduction}
Over the past few decades, robotic manipulators have evolved into highly sophisticated systems with diverse morphologies, which have been deployed to accomplish various tasks, including pick-and-place~\citep{kim1987visual, wang2021optimal}, welding~\citep{ogbemhe2015towards, kah2015robotic}, polishing~\citep{xu2017kinematics, kharidege2017practical}, and human-robot collaboration~\citep{murphy2010human, goodrich2008human}. Although effective in specific applications, these manipulators are built with fixed morphologies, limiting their adaptability to varying scenarios. To address this, modular manipulators were introduced~\citep{matsumaru1995design, zhang2006novel, yim2007modular, romiti2021toward, rossini2025concert}. 
Made of interchangeable, pre-manufactured basic modules, modular manipulators can be rapidly assembled into different morphologies, facilitating quick deployment across a wide range of scenarios. However, adapting modular manipulators to complex, task-specific requirements remains challenging, necessitating an effective computational design framework that simultaneously addresses task-driven, feasible motion planning and the co-optimization of both morphology and mounted pose.

\begin{figure}[h]
  \centering
\includegraphics[width=0.85\linewidth]{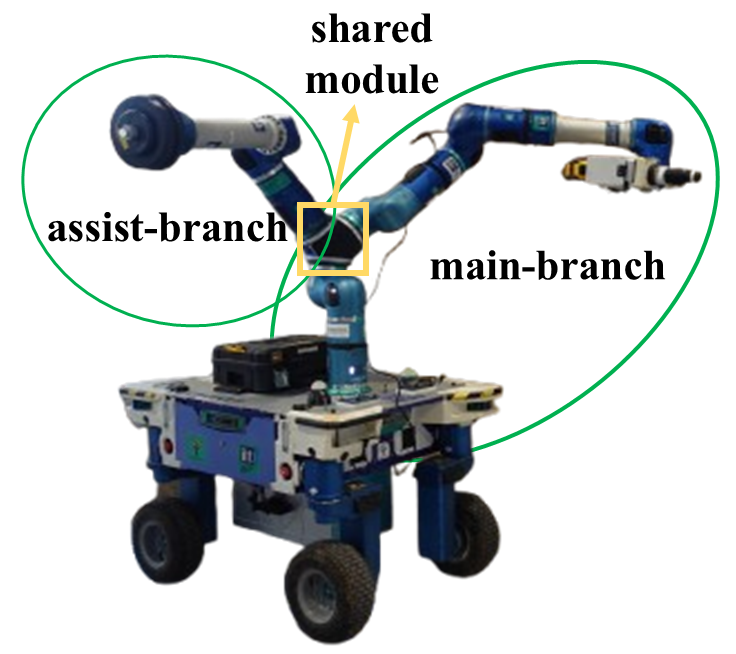}\vspace{-2mm}
  \caption{Concept of a bi-branch modular manipulator consisting of a main branch with an end-effector and an assist branch that is connected via a shared module.
}
  \label{fig:concept}
\end{figure}

A manipulation task is typically defined by a reference trajectory at the end effector~\citep{shiller1985optimal}, necessitating precise control of both translational and rotational movements. Depending on the number of degrees of freedom (DoF) and task requirements, a manipulator can be classified as redundant or non-redundant. 
By providing additional DoFs, redundant morphologies enable the concurrent fulfilment of multiple sub-tasks at the end-effector, which, however, may need extra cost. In contrast, although non-redundant morphologies are more compact and cost-friendly, they often encounter conflicts when attempting to execute desired trajectories that demand full control of both position and orientation, due to the lack of DoFs. However, realizing the unified motion planner across the different morphologies to adapt to various tasks is challenging. In addition, modular manipulators also require the selection of morphology and mounted pose to meet task and performance requirements.  
For morphological design optimization\footnote{Morphological design refers to the joint optimization of a manipulator’s morphology and its mounted pose.}, the optimal morphology resides in a discrete space, while the mounted pose is adjusted within a continuous space. This hybrid property poses a significant challenge for the co-optimization of morphology and mounted pose.


To address the above challenges, several studies have been reported. For example, the existing work in \citep{ha2018computational,romiti2021minimum, lei2024task, mayer2025holistic} integrated general-purpose planners with a unified morphological design, allowing task-driven co-adaptation of motion and design for modular manipulators. However, the motion planner in these frameworks only works for the redundant manipulator.
Furthermore, these frameworks are primarily designed for single-branch robots. When working with a large workspace, the single-branch morphology often leads to high proximal joint torque due to the long-reaching structure. Since the module’s torque capacity is fixed once manufactured, the above weak-point limits the use of single-branch designs in high-load scenarios. In contrast, multi-branch manipulators with assistive branches show better performance in handling complex tasks and heavier payloads~\citep{lei2022dual, kennel2024payload, raina2021impact}, as the additional branch helps redistribute loads and support the main-branch's motion. Nevertheless, such structures significantly increase design dimensionality and planning complexity. Until now, there is still a lack of a generalizable and efficient design framework that can handle both single-branch and multi-branch cases.

In this work, we propose a {planner-in-the-loop} computational design framework for modular manipulators that unifies motion planning and design optimization. 
Unlike previous works~\citep{lei2024task, kulz2024optimizing}, which focused only on non-redundant morphologies, the proposed framework handles both the redundant and non-redundant cases by introducing a hierarchical model predictive control (HMPC)-based planner. By formulating redundancy at the task level, HMPC enables non-redundant manipulators to handle multiple tasks without inducing conflicts. The planned trajectory is also used to evaluate each candidate design with task-specific performance metrics, tightly coupling motion planning and design optimization. Then, we adopt a sorting-based mapping function to transform the discrete module selection and arrangement into a continuous search problem, allowing us to use an efficient search algorithm such as CMA-ES~\citep{krause2016cma} to optimize morphology and mounted pose simultaneously. To ensure a feasible configuration, we impose physical constraints such as those on tracking accuracy and dynamic feasibility during the design optimization process. Furthermore, this framework allows for customizing the manipulator's design with various objectives, such as maximizing manipulability, minimizing joint effort, and reducing module usage. 

In addition, we introduce a bi-branch morphology (see Fig.~\ref{fig:concept}) to enhance the capability of modular manipulators, without requiring actuator upgrades or the re-manufacturing of basic modules. 
In this bi-branch structure, the main branch performs the task, while the assistive branch counteracts dynamic disturbances induced by the main branch and enables load redistribution and torque reduction at the proximal joints. To support both single-branch and bi-branch optimization in a unified manner, we extend the morphology representation by incorporating a virtual module that serves as a segmentation marker. 
In this way, both single-branch and bi-branch morphologies can be described in a chain-type format. {Compared with the existing work}, this design framework can automatically select between single-branch and bi-branch morphologies according to task requirements.

The main contributions of this work include: 
\begin{itemize}
\item We develop a unified, {planner-in-the-loop} design framework that integrates motion planning and design optimization. The novel motion planner enables us to handle both redundant and non-redundant morphologies. The {planned trajectory is also evaluated for iteratively co-optimizing} morphology and mounted pose. 
\item We introduce the concept of a bi-branch manipulator, where an assistive branch is integrated into a traditional single-branch structure. 
Through introducing the virtual module in the sorting mapping function and incorporating the motion planning for the assisted branch, our framework can automatically select between single-branch and bi-branch structures. To the best of our knowledge, this is among the first to explore such functional modular manipulators. 
\item We conduct extensive simulations and experiments across three representative task scenarios, demonstrating the effectiveness of the proposed method under varying task objectives. Comparison studies demonstrate superiority over other baseline methods.
\end{itemize}

{
This work substantially extends our previous study~\citep{lei2024task}. The main differences are (i) We propose a unified HMPC-based motion planner for modular manipulators, applicable to both non-redundant and redundant morphologies.
(ii) We introduce a bi-branch morphology with an integrated assistive branch to enhance single-branch capabilities, while proposing a computational design framework that supports the optimization and selection of both single-branch and bi-branch morphologies.   
(iii) We conduct extensive simulations and hardware experiments to fully validate the effectiveness of the proposed framework.
}

The remainder of this article is organized as follows. Sec.~\ref{related_work} surveys related work. 
Sec.~\ref{framework_overview} provides a first glance on the proposed framework. 
Sec.~\ref{hardwareadntasktrajectroy} introduces the preliminary knowledge regarding the modular robotic system and the trajectory generation methodology. Sec.~\ref{sec:HMPC} details the HMPC-based planner. Sec.~\ref{design_optimization} outlines the structure optimization approach. Sec.~\ref{experimentandsimulation} evaluates the proposed method via simulation and hardware experiments. Finally, Sec.~\ref{conclusion} concludes this paper.

\section{Related work}~\label{related_work}
\textcolor{blue}{}




Computational design frameworks play a central role in the manufacture and application of modular manipulators. Our contribution lies in integrating an MPC-based motion planner with a planner-in-the-loop design optimization method to determine task-specific morphologies and motion trajectories. The following survey focuses on two key components: the MPC-based motion planner and the optimal design methods for modular manipulators.


\subsection{MPC-based Planner for Redundant and Non-redundant Manipulator}
The motion planner in the computational design framework is designed to generate feasible motion trajectories across various morphologies and guide optimal morphological design. To achieve this, previous computational design frameworks for modular robots~\citep{zhao2020robogrammar, lei2024task} have adopted MPC-based planners, which formulate an optimization problem over a receding time horizon to predict the future state of the system and adjust control inputs accordingly, while ensuring compliance with the dynamics and constraints of the system~\citep{ qin1997overview, mayne2014model, kohler2018nonlinear}.
In addition to motion generation, MPC-based planners have also been extended to ensure successful execution by mitigating the adverse effects of singular configurations through predictive control~\citep{wang2024hierarchical, lee2023real}, and to support deployment in dynamic environments by incorporating collision avoidance constraints into the MPC formulation~\citep{nubert2020safe, lei2022mpc, gafur2021dynamic, gaertner2021collision, kramer2020model}. However, these MPC-based planners have not been generalized to apply to both redundant and non-redundant manipulators.

Although some manipulators are not redundant, they can still benefit from redundancy-based planning strategies~\citep{slotine1991general,mansard2009directional}. This is because, in many practical applications, full tracking in all task-space directions is unnecessary. For instance, tasks such as arc welding~\citep{huo2005kinematic} or spray painting~\citep{zanchettin2011use} can often be performed using 5- or 6-DoF manipulators without requiring orientation tracking along the $z$-axis. When the DoFs exceed the dimensionality of the primary subtasks, the system demonstrates \textit{functional redundancy}~\citep{nicolis2020operational}, which enables more flexible and conflict-resilient planning~\citep{siciliano1990kinematic, mansard2009directional}. 
To support such prioritization, hierarchical MPC frameworks have been proposed~\citep{minniti2019whole, bouyarmane2017weight}, assigning higher weights to critical tasks and lower weights to secondary ones. However, the above frameworks require manual weight tuning when task requirements or the manipulator's morphology change, which poses a particularly pronounced issue for unifying the planner with the non-redundant and redundant morphologies. Until now, a unified MPC-based planner for modular manipulators that enhances adaptability without the need for further parameter adjustments is still missing.

\subsection{Design Optimization for Modular Manipulators} 
Morphology optimization entails selecting and arranging modular components to construct a manipulator capable of accomplishing a specific task. To solve this discrete combinatorial problem, a straightforward strategy is to enumerate all feasible morphologies~\citep{liu2020optimizing, romiti2021minimum, sathuluri2023robust}, which ensures completeness but becomes computationally intractable as the number of modules increases. Instead, heuristic methods have been widely adopted to improve search efficiency within the discrete design space~\citep{ha2018computational, kulz2024optimizing, icer2017evolutionary, zhao2020robogrammar, koike2023simultaneous, hu2023glso}. While these methods are effective in identifying functional morphological configurations, they often overlook the optimization of the manipulator’s mounted pose\footnote{Mounted pose refers to the placement and orientation of the entire structure within the task environment}. The mounted pose optimization, defined in a continuous space, plays a critical role in enhancing the reachability, motion precision, and task feasibility~\citep{cursi2022optimization, qin2022install, du2024learning}. Since both morphology and mounted pose collectively determine the performance, optimizing them independently may result in feasible designs that are structurally sound but poorly adapted for task execution,  highlighting the importance of concurrent optimization of morphology and the mounted pose.

Further observations reveal that the concurrent optimization of morphology and mounted pose requires searching in a hybrid discrete and continuous space, which is a challenging task. To solve this problem, prior works~\citep{mayer2025holistic, romiti2021minimum} proposed discretizing the continuous pose space into a finite set of candidate placements, thereby reformulating the co-optimization problem as a purely combinatorial one. While this approach facilitates simultaneous optimization, it risks excluding high-performing solutions due to the limited resolution of the discretized space. In contrast, our previous work~\citep{lei2024task} introduced a sorting-based mapping function that encodes discrete morphologies into a continuous domain.
This formulation supports co-optimization of morphology and pose within a unified continuous space, alleviating the suboptimality introduced by discretization. 

Although these methods~\citep{mayer2025holistic, romiti2023optimization, lei2024task} offer promising solutions for co-optimizing structure and placement, they remain constrained to single-branch morphologies. {Differing from the single-branch morphology, multi-branch manipulators have demonstrated potential for performing complex tasks by enabling greater payload capacity and extended end-effector workspace~\citep{kennel2024payload, romiti2021minimum, lei2022dual, whitman2018task, du2024learning}. However, the existing works do not address the optimization of the morphology for multi-branch manipulators within the task context. Furthermore, the mounted pose of the multi-branch manipulator is not optimized yet.}

\begin{figure*}
  \centering
\includegraphics[width=0.99\textwidth]{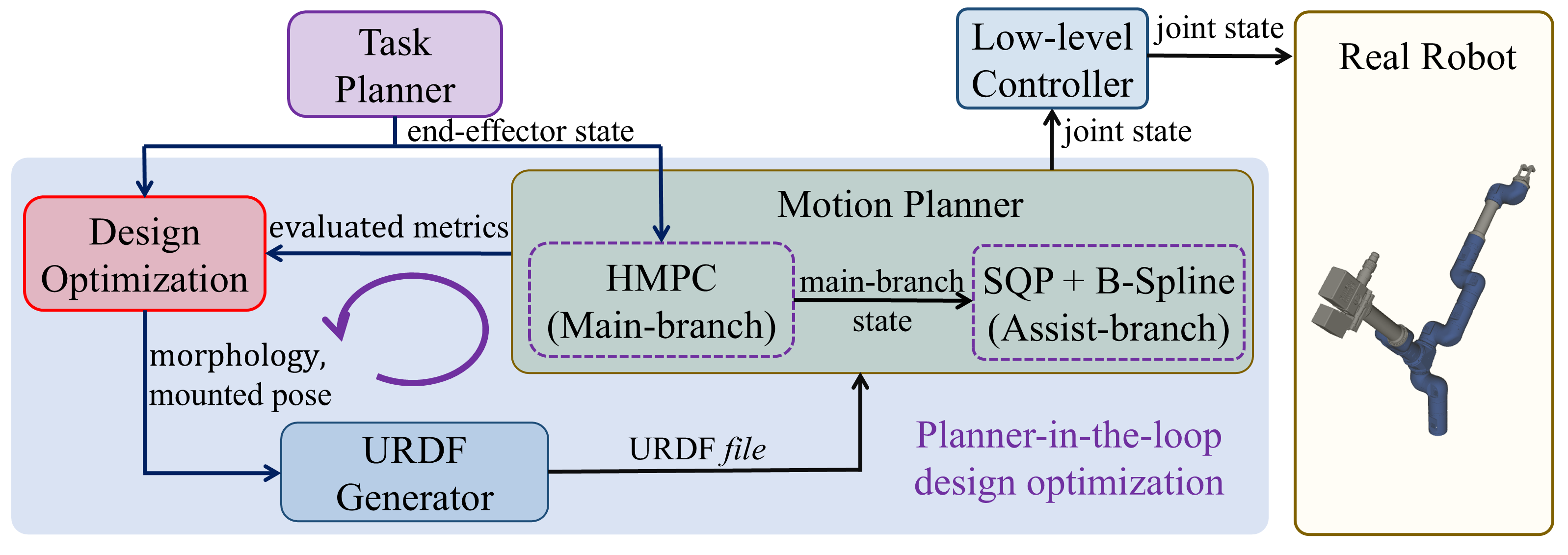}\vspace{-2mm}
  \caption{\textbf{Schematic of the Computational Design Framework}. The manipulation task is defined as a sequence of end-effector trajectories (including position and orientation) in Cartesian space. For the main branch, we employ an HMPC-based planner to compute feasible joint-space trajectories. For the assist branch (in the bi-branch morphology), an additional planning component is introduced to determine the joint movement. With the motion planner in the design loop, the execution performance—measured using designated evaluation metrics—is maximized to refine both the manipulator's morphology and its mounted pose. Note that this framework can automatically select between single-branch and bi-branch morphologies according to the task requirements.}
  \label{fig:framework}
\end{figure*}

\section{Framework Overview}~\label{framework_overview}
Fig.~\ref{fig:framework} illustrates the overall computational framework for designing the modular manipulator. The framework comprises three main components: the task planner, which defines the end-effector task requirements; the motion planner, which ensures effective task execution; and the design optimization module, which selects the manipulator's morphology and determines the mounted pose.

In this framework, the task was specified by the end-effector reference trajectory. In the motion-planning component, the HMPC generates feasible joint trajectories, giving different single-branch morphologies. For bi-branch morphologies, the motion planner will further identify and optimize the joint motion of the assist branch. The resultant movement trajectories serve two purposes: (i) they are executed by low-level controllers in simulation or hardware experiments to achieve the desired task, and (ii) they are fed back into the design optimization loop, refining the design of the manipulator.

In the design optimization process, the trajectories generated by the motion planner are evaluated using task-specific performance metrics, which are then input to the design optimizer. These evaluated metrics serve as heuristic objectives to guide the refinement of the morphological design. This process is formulated as an in-loop optimization framework, where the motion planner and the design optimizer interact iteratively, thus improving the adaptability to various task requirements.






\section{Task Planning}~\label{hardwareadntasktrajectroy}
\subsection{Hardware Description}~\label{hardware}
The basic modules used in this work are those comprising the CONCERT robot~\citep{rossini2025concert}. For a thorough hardware description of the modules, as well as details on their interconnections and connection capabilities, the reader can refer to the cited paper. In summary, the set of modules includes:

\textit{(i)} Straight-Joint module: introduces a rotation about the yaw axis that is parallel to the shared central normal of the modular input and output flanges. In the current work, two sizes are provided, with the maximal continuous torque being \qty{120}{\newton\meter} and \qty{160}{\newton\meter}, respectively. 

\textit{(ii)} Elbow-Joint module: generates rotation about the pitch axis that is perpendicular to the central normal. Two sizes with the same torque capabilities as the `Straight-Joint' modules are offered.

\textit{(iii)} Passive-Link module: motor-less links with three lengths—\qty{0.3}{\meter}, \qty{0.4}{\meter}, and \qty{0.6}{\meter} to extend the reachability.

\textit{(iv)} “Y” module: splits one chain into two branches. As shown in Fig.~\ref{fig:concept}, the “Y” module serves as a shared connection between the main branch and the assist branch.

Detailed descriptions of the above modules are illustrated in Fig.~\ref{fig:representation} (left), where \( \{ f_{\text{in}}^i \} \) and \( \{ f_{\text{out}}^i \} \) represent the input and output connectors of the \(i\)-th module, respectively. The input connector refers to the interface connecting to the previous module in the chain, typically closer to the base link, while the output connector connects to the subsequent module. Additionally, \( \{ f_\text{joint}^i \} \) represents the motion position of the motors in each joint module. Notably, the  “Y” module includes two different output connections.





\subsection{End-effector Reference Trajectory}~\label{task_trajectory_gen}
The desired task is defined as the end-effector trajectory in Cartesian space, represented by a sequence of desired poses to be tracked at each time step. Such trajectories are determined by defining waypoints that the end-effector must pass through.

\subsubsection{Position trajectory generation}

The position trajectory is generated by the \(n\)-th order polynomial interpolation:
\begin{equation}
\bm{p}(t) = \sum_{i=0}^{n} \bm{a}_i t^i,
\end{equation}
where \(\bm{p}(t) \in \mathbb{R}^3\) denotes the end-effector position at time \(t\), and \(\bm{a}_i\) are the polynomial coefficients to be optimized. 

To satisfy multiple requirements such as smoothness, dynamic feasibility, and collision avoidance, the optimal trajectory is generated by solving:

\begin{equation}
\begin{aligned}
\min_{\bm{a}_i} \quad & \int_{0}^{T} \left\| \frac{d^2 \bm{p}(t)}{dt^2} \right\|^2 dt \quad  \\
\text{s.t.} \quad
& \qquad \!\!\! \bm{p}(t_k) = \bm{p}_k^{\text{des}}, \qquad k = 1, \dots, K , \\
& \left\| \frac{d \bm{p}(t)}{dt} \right\| \leq v_{\max}, \quad \forall t \in [0, T], \\
& \left\| \! \frac{d^2 \bm{p}(t)}{dt^2} \right\|  \! \leq a_{\max}, \quad \! \forall t \in [0, T] \quad \\
& \bm{p}(t) \notin \mathcal{O}, \quad \forall t \in [0, T] ,
\end{aligned}
\label{position_trajectory}
\end{equation}
where \(\{\bm{p}_k^{\text{des}}\}\) are the desired waypoints in the world frame($t_k$ represent the corresponding time moments), and \(\mathcal{O}\) represents the obstacle region in the workspace.


\subsubsection{Orientation trajectory generation}
The orientation trajectory is also generated by interpolating among predefined waypoints. In particular, a smooth transition is required between each pair of adjacent orientation waypoints. For each segment, a continuous interpolation is performed from the initial orientation (represented by the rotation matrix \(\mathbf{R}_{\text{in}}\in \mathbb{R}^{3\times3}\)) to the desired orientation (\(\mathbf{R}_d\in \mathbb{R}^{3\times3}\)) using the exponential map formulation:
\begin{equation}
\mathbf{R}(t) = \mathbf{R}_{\text{in}} \exp\left( \frac{t - t_k}{t_{k+1} - t_k} \log\left( \mathbf{R}_{\text{in}}^{-1} \mathbf{R_d} \right) \right),
~\label{orinetation_trajectory}
\end{equation}
where \(t \in [t_k, t_{k+1}]\) denotes the time interval between two consecutive waypoints and \(\mathbf{R}(t)\) represents the interpolated rotation matrix at time \(t\). The expression \(\log\left( \mathbf{R}_{\text{in}}^{-1} \mathbf{R_d} \right)\) gives the corresponding rotation vector in the lie algebra \(SO(3)\) manifold, which is then proportionally scaled over time using the exponential map. 

The above position and orientation trajectories exhibit differentiable continuity, ensuring smooth and dynamically feasible motion of the end-effector. Moreover, the time-aligned formulation guarantees that both position and orientation evolve synchronously, thereby enabling coherent and consistent task-space motion planning, which will be detailed in the next section.

\section{Take-driven Motion Planning}~\label{sec:HMPC}
This section introduces the motion planner, handling both the single-branch and the bi-branch cases. In the single-branch case, the planner optimizes joint motions to execute the desired Cartesian motion of the end-effector. In the bi-branch case with an additional assistive branch, the planner optimizes the main-branch motion to follow the reference trajectory while simultaneously coordinating the assist branch to reduce loads on the proximal-base joints.

\subsection{HMPC based Planner for Task Execution}~\label{MPC_formulation}

In the MPC-based planner proposed in the prior MPC formulations of the computational framework~\citep{lei2024task}, the reference task \(\mathcal{T}\) is defined as 6D end-effector tracking (position and orientation along the \(x\), \(y\), and \(z\) axes), i.e., \(\operatorname{dim}(\mathcal{T}) = 6\). For redundant manipulators with its DoFs \(\operatorname{dim}(\mathcal{R}) > \operatorname{dim}(\mathcal{T})\), such tasks can be executed by exploiting the additional DoFs. In contrast, non-redundant manipulators lack sufficient DoFs, which limits their dexterous workspace~\citep{gupta1986nature} and makes full-task execution more challenging due to potential joint-space conflicts under the same MPC formulation.



To unify the motion generation across both redundant and non-redundant morphologies, we adopt a two-level HMPC strategy, where the high-level MPC plans joint motions to accomplish high-priority sub-tasks, and the low-level MPC refines the execution of lower-priority sub-tasks. 

\subsubsection{Kinematics model}
We define the end-effector state as
\(
\bm{x}_e := [\bm{p}^T\ \! \ \bm{o}^T \ \! \ \bm{\dot{p}}^T \ \ \! \bm{\omega}^T ]^T \in \mathbb{R}^{13},
\)
where \(\bm{p} \in \mathbb{R}^{3}\) and \(\bm{\dot{p}} \in \mathbb{R}^{3}\) denote the end-effector position and linear velocity in the world frame \(\{\bm{W}\}\), respectively. The quaternion \(\bm{o}=[\eta, \bm{\epsilon}^T]\in \mathbb{R}^{4}\) represents the end-effector's orientation relative to \(\{\bm{W}\}\), with \(\|\bm{o}\|=1\), \(\eta=o_w\) and \(\bm{\epsilon}=[o_x,o_y,o_z]^T\). \(\bm{\omega} \in \mathbb{R}^{3}\) represents the global angular velocity. 

The relationship between the joint motion and end-effector motion is characterized using the Jacobian matrix \(\mathbf{J} \in \mathbb{R}^{6 \times n_j}\), where \(n_j\) is the number of joints. This matrix, along with its time derivative \(\dot{\mathbf{J}} \in \mathbb{R}^{6 \times n_j}\), enables the transformation from joint space to end-effector space:
\begin{equation}
\dot{\bm{x}}_e := \!\!\!\!\!\!\!\!\!\!\!
\underbrace{
\begin{bmatrix}
\bm{v}_e \\
\dot{\bm{v}}_e
\end{bmatrix}
}_{\substack{
\text{End-effector kinematics} \\
\dot{\bm{x}}_e \in \mathbb{R}^{12}
}} \!\!\!\!\!\!\!\!
= 
\underbrace{
\begin{bmatrix}
\mathbf{J}(\bm{q})  & \bm{0}_{6 \times n_j}\\[3pt]
\dot{\mathbf{J}}(\bm{q},\bm{\dot{q}}) & \mathbf{J}(\bm{q})
\end{bmatrix}
}_{\substack{
\text{Kinematic mapping} \\
\mathbf{B}_{\text{kin}} \in \mathbb{R}^{12 \times 2n_j}
}}
\ \!\!\!\!\!\!\!\!
\underbrace{
\begin{bmatrix}
\dot{\bm{q}} \\[3pt]
\ddot{\bm{q}}
\end{bmatrix}
}_{\substack{
\text{Joint velocity and} \\
\text{acceleration} \ \ \bm{u}_{\text{kin}} \in \mathbb{R}^{2n_j}
}}
\label{eq:kinematics_model}
\end{equation}
where \(\dot{\bm{q}} \in \mathbb{R}^{n_j}\) and \(\ddot{\bm{q}} \in \mathbb{R}^{n_j}\) are the joint velocity and acceleration, respectively; \(\bm{v}_e = [\bm{\dot{p}}^T \ \ \bm{\omega}^T]^T\in \mathbb{R}^6\) and \(\dot{\bm{v}}_e = [\bm{\ddot{p}}^T \ \ \bm{\dot \omega}^T]^T \in \mathbb{R}^6\) represent the end effector velocity and acceleration, respectively. 


\subsubsection{State-space formulation}

The derivation of \(\bm{x}_e\) is
\(
\bm{\dot{x}}_e := [\bm{\dot{p}}^T\ \ \bm{\omega}^T  \ \ \bm{\dot{v}}^T \ \ \bm{\dot{\omega}}^T ]^T \in \mathbb{R}^{12}
\).
Specifically, the relationship between the angular velocity \(\bm{\omega}\) and the derivative of the quaternion \(\bm{\dot{o}}\) at the \(k\)-th step is defined with
\(
\dot{\bm{o}}(t) = \frac{1}{2} \mathbf{G}(\bm{o}_k) \bm{\omega}_k
\)
where \(\mathbf{G}(\bm{o}) \in \mathbb{R}^{4\times3} \) is defined as
\(
\mathbf{G}(\bm{o}) = \begin{bmatrix}
-\bm{\epsilon}^\top, 
(\eta \bm{I}_3 + \hat{\bm{\epsilon})^T}
\end{bmatrix},
\)
in which \(\hat{\bm{\epsilon}}\) is the skew-symmetric matrix constructed from \(\bm{\epsilon}\), and \(\eta\) is the scalar part. With this model, the state-space equation of the end-effector, following the forward Euler integration with time step \(dt\), is
\begin{equation}
\begin{bmatrix}\!
\bm{p}_{k+1} \\
\bm{o}_{k+1} \\
\bm{v}_{k+1}\\
\bm{\omega}_{k+1}\!\!
\end{bmatrix} \!\!\!=\!\! \!
\begin{bmatrix}\!
\bm{p}_k \\
\bm{o}_k \\
\bm{v}_k \\
\bm{\omega}_k\!
\end{bmatrix}
\!\!+\!\!
\underbrace{\begin{bmatrix}\!
\mathbf{I}dt & 0 & 0 & 0 \\
0 & \frac{1}{2} \textbf{G}(\bm{o}_k)dt & 0 & 0 \\
0 & 0 & \mathbf{I}dt & 0 \\
0 & 0 & 0 & \mathbf{I}dt \!
\end{bmatrix}}_{\mathbf{\bm{B}}_{\text{e,k}}}
\!\!\begin{bmatrix}\!
\dot{{p}}_k \\
\boldsymbol{\omega}_k \\
\dot{{v}}_k \\
\dot{\boldsymbol{\omega}}_k\!
\end{bmatrix}.
\end{equation}

Considering Eq.~\eqref{eq:kinematics_model}, we can define the predictive model as 
\begin{equation}
\bm{{x}}_{e,k+1} = \bm{{x}}_{e,k} +
\mathbf{\bm{B}}_{\text{e,k}} \mathbf{\bm{B}}_{\text{kin,k}} \bm{u}_{\text{kin,k}}.
\label{MPC_kinematicals_discrete}
\end{equation}

\subsubsection{Two-level MPC} 
{To realize hierarchical motion planning, we decompose the end-effector task into six directional components, assigning them various priority levels as either high or low.} The HMPC-based planner operates in the hierarchical layers: the \textbf{high-level MPC} generates state trajectories fulfilling the high-priority sub-tasks (\( \mathcal{T}_H \)) requirements, while the \textbf{low-level MPC} subsequently addresses the low-priority sub-tasks (\( \mathcal{T}_L \)), taking the high-priority trajectories as hard constraints.

\textbf{High-level MPC formulation}: The high-level MPC computes the optimal joint velocities and accelerations to achieve high-priority tasks. 
Choosing the high-priority control input as decision variable \( \bm{u}_{\text{kin},k}^{(1)}\), the high-level MPC is formulated as
\begin{subequations}
\begin{align}
    \min_{\bm{u}_{\text{kin}}^{(1)}} \!\!\!\!\! \quad & \sum_{k=1}^{N_h} \left( 
        \left\| \bm{x}_{e,k+1}^{(1)} - \bm{x}_{e,\text{ref},k+1}^{(1)} \right\|_{\mathbf{Q}_k^{(1)}}^2 
       \! + \! \left\| \bm{u}_{\text{kin},k}^{(1)} \right\|_{\mathbf{R}_k}^2 
    \right) \\
    \text{s.t.} \quad & 
    \bm{x}_{e,k+1}^{(1)} = \bm{x}_{e,k}^{(1)} + \mathbf{B}_{\text{e},k} \mathbf{B}_{\text{kin},k}(\bm{q}_1,\dot{\bm{q}}_1) \bm{u}_{\text{kin},k}^{(1)} \\
    & \bm{q}_l \leq \mathbf{E} \, dt \, \dot{\bm{q}}_k + \bm{q}_0 \leq \bm{q}_u ~\label{eq:joint_position_constraint}
\\
    & \dot{\bm{q}}_{\min} \leq \dot{\bm{q}}_k \leq \dot{\bm{q}}_{\max} \\
    & \ddot{\bm{q}}_{\min} \leq \ddot{\bm{q}}_k \leq \ddot{\bm{q}}_{\max}
\end{align}
 \label{eq:highlevel_MPC}
\end{subequations}
where the superscript \( (1) \) indicates the association with high priority tasks, \(N_h\) is the prediction horizon, \( \mathbf{Q}_k^{(1)} \) defines the weighting matrix for high priority tasks, \( \mathbf{R}_k \) defines the weight of the regularization term.  Eq.~\eqref{eq:joint_position_constraint} represents the joint position constraints at each joint. Specifically, the joint position \(
\bm{q}_k \in \mathbb{R} ^{n_j} \) in the $k$ -th step can be approximated as
\(
\bm{q}_k \approx \bm{q}_0 + \sum_{m=1}^{k} \dot{\bm{q}}_m \,dt
\). Given the joint position bounds \(\bm{q}_l \leq \bm{q}_k \leq \bm{q}_u\), this joint limit can be expressed as a linear form in Eq.~\eqref{eq:joint_position_constraint}, with \( \mathbf{E} \) representing the lower-triangular all-one matrix.

The high-level MPC is formulated in analogy to the standard MPC framework~\citep{holkar2010overview} to accomplish high-priority subtasks. In this formulation, the state prediction function in Eq.~\eqref{MPC_kinematicals_discrete} is linearized under the assumption that the Jacobian $\bm{\mathbf{J}_{v}}$ and its time derivative $\bm{\mathbf{\dot{J}}_{v}}$ (see Eq.~\eqref{eq:kinematics_model}) remain constant within the prediction horizon, allowing the high-level MPC problem to be solved via quadratic programming (QP). 





\textbf{Low‐level MPC formulation:}  The low-level MPC {aims to} refine the joint-space trajectory to accommodate secondary subtasks by leveraging the remaining DoFs, while treating the high-level MPC outputs as hard constraints. Specifically, the optimal end-effector states predicted by the high-level MPC are imposed as fixed constraints in the low-level MPC. The low-level MPC can be formed as
\begin{subequations}
\begin{align}
\min_{\bm{u}_{\text{kin}}^{(2)}} \!\!\!\!\! \quad & \sum_{k=1}^{N_l} \left( 
\left\| \bm{x}_{e,k+1}^{(2)} \!-\! \bm{x}_{e,\text{ref},k+1}^{(2)} \right\|_{\mathbf{\mathbf{Q}_k^{(2)}}}^2 \!+\! 
\left\| \bm{u}_{\text{kin},k}^{(2)} \right\|_{\mathbf{R_k}}^2 
\right) \\
\text{s.t.} \quad 
& \bm{x}_{e,k+1}^{(2)} = \bm{x}_{e,k}^{(2)} + 
\mathbf{\mathbf{B}_{\text{e},k}} \mathbf{\mathbf{B}_{\text{kin},k}}\bigl(\bm{q}_{k}, \bm{\dot{q}}_{k}\bigr) 
\bm{u}_{\text{kin},k}^{(2)}, \\
& \bm{q}_l \leq \mathbf{E} \, dt \, \bm{\dot{q}}_k + \bm{q}_0 \leq \bm{q}_u, 
\quad \forall k = 1, \ldots, N_l, \\
& \dot{\bm{q}}_{\min} \leq \bm{\dot{q}}_k \leq \dot{\bm{q}}_{\max}, 
\quad \forall k = 1, \ldots, N_l, ~\label{constraints_state}\\
& \ddot{\bm{q}}_{\min} \leq \bm{\ddot{q}}_k \leq \ddot{\bm{q}}_{\max}, 
\quad \forall k = 1, \ldots, N_l, \\
& \mathbf{\mathbf{B}_{\text{kin},k}}\bigl(\bm{q}_{k}, \bm{\dot{q}}_{k}\bigr) 
\bm{u}_{\text{kin},k}^{(2)} = 
\mathbf{\mathbf{B}_{\text{kin},k}}\bigl(\bm{q}_1, \bm{\dot{q}}_1\bigr) 
\bm{u}_{\text{kin},k}^{(1)}, \nonumber \\
& \qquad \forall k = 1, \ldots, N_l, ~\label{eq:state_constraint}
\end{align}
\label{eq:low_level_MPC}
\end{subequations}
where the superscript \( (2) \) indicates association with the secondary tasks and \(N_l\) represents the prediction horizon. In Eq.~\eqref{eq:state_constraint}, the constraints are formulated to ensure that the end-effector consistently executes the high-priority sub-task, using the kinematic model that maps the joint state \(\bm{u}_{\text{kin}}^{(1)}\) to the corresponding end-effector state. Under these constraints, we achieve lower-priority sub-tasks without compromising the high-priority sub-tasks.

In both high-level and low-level MPC formulations, kinematic constraints are enforced, including joint position, velocity, and acceleration limits. However, dynamic constraints are ignored, which will be addressed during the design optimization phase, as discussed in the Sec.~\ref {sec:dyn_model} and~\ref{optimalcames}. Meanwhile, to ensure that the HMPC can yield feasible and optimal joint states, the hierarchical MPC structure should maintain a redundancy-based formulation. Specifically, the input dimension (i.e., DoF number) must exceed the output dimension (i.e., number of task-space constraints at each priority level). 
Thus, it becomes necessary to ensure that the number of DoFs exceeds the number of task-space constraints at each priority level. This condition allows the system to be formulated redundantly, even if it is not structurally redundant. To meet this requirement, the total number of DoFs must be explicitly optimized and constrained, which will be described in Sect.~\ref{optimalcames}.




\subsection{ Assist Branch Trajectory Optimization}
\begin{figure}
  \centering
\includegraphics[width=01.0\linewidth]{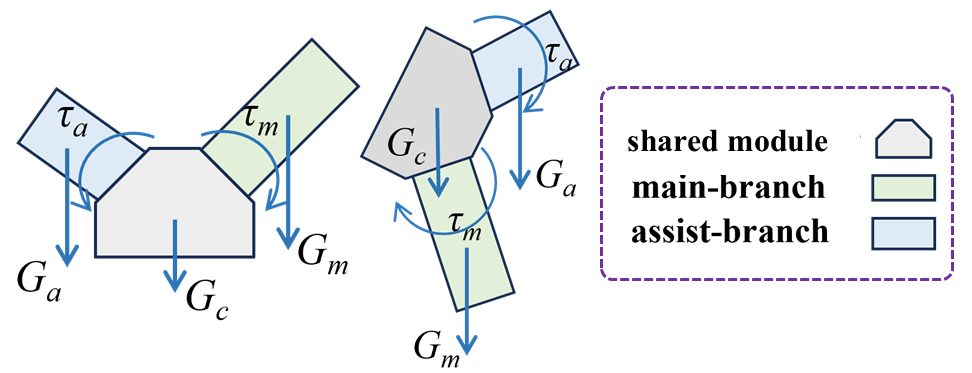}\vspace{-2mm}
  \caption{\textbf{Comparison of gravitational torque under two different CoM placements}. Left scenario: CoMs of both branches lie on opposite sides of the shared module, producing a balancing effect. Right scenario: CoMs are on the same side, leading to increased gravitational torque.}
  \label{fig:balance_updae}
\end{figure}

For bi-branch morphologies, the torque mitigation depends on the spatial arrangement of the two branches. As shown in Fig.~\ref{fig:balance_updae}, when the centers of mass (CoMs) are positioned on opposite sides of the shared module (left subfigure), the gravitational torque from the assist branch (\(\tau_a\))  partially offsets that of the main branch (\(\tau_m\)), reducing the load on the proximal joints. In contrast, when both CoMs lie on the same side, the torques compound, increasing the demand at the base. 
{Therefore, the assist branch should dynamically adjust its state, ensuring effective load compensation during the task execution process.}

The reference trajectory of the bi-branch manipulator is defined by the end-effector motion of the main branch, following the HMPC trajectory.
Corresponding, the joint movement of the assist branch is optimized by solving
\begin{equation}
\begin{aligned}
\min_{\mathbf{q}, \dot{\mathbf{q}}, \ddot{\mathbf{q}}} \quad 
& \sum_{i=1}^{n} \|\mathbf{\tau}_i\|^2 
+ \lambda \sum_{i=1}^{n} ({q}_i - {q}_{0,i})^2 \\
& + \underbrace{\mu \sum_{i=1}^{n} \left( \frac{1}{\alpha} \ln\bigl(1 + e^{\alpha (\|\mathbf{\tau}_i\| - \tau_{\max,i})}\bigr) \right)^2}_{\text{penalize excessive joint torque limitation}} \\
\text{s.t.} \quad 
& \underbrace{\mathbf{M}(\bm{q}) \ddot{\bm{q}} 
    \!+\! \mathbf{C}(\bm{q}, \dot{\bm{q}}) \dot{\bm{q}} 
    \!+\! \mathbf{g}(\bm{q}) 
   \! +\! \mathbf{h}_{\text{payload}}(\bm{q}) 
    \!=\! \bm{\tau}}_{\text{full-body dynamics}},\\
& \underbrace{\bm{q}_{m,i} = \bm{q}_{d,i}}_{\text{main-branch state constraints}},\\
& \bm{q}_{\min,i} \leq \bm{q}_{i} \leq \bm{q}_{\max,i}, \\
& \bm{\dot{q}}_{\min,i} \leq \bm{\dot{q}}_{i} \leq \bm{\dot{q}}_{\max,i}.
\end{aligned}
\label{Eq:SQP}
\end{equation}
where \(\bm{q}, \dot{\bm{q}}, \ddot{\bm{q}} \in \mathbb{R}^{d_a} \) are the joint positions, velocities, and accelerations of the assist branch, with $d_a$ being the DoFs of the assist branch; \(\bm{q}_{m,i} \in \mathbb{R}^{d_m}\) and \(\bm{q}_{d,m} \in \mathbb{R}^{d_m}\) represent the main-branch's current and desired joint position at \(i\)-th joint, with $d_m$ being the DoFs of the main branch. 

The objective function aims to minimize the overall joint effort of the entire manipulator while keeping the solution close to the robot’s initial joint state $\bm{q}_{0}$. The first equality constraint enforces the full-body dynamics, and the second preserves the main-branch states determined by HMPC. Since non-linear constraints exist, such as the full-body dynamics, the above optimization problem is a general non-linear programming problem, which is then solved via sequential quadratic programming (SQP)~\citep{gill2005snopt}. 



The joint motions of the assistive branch, obtained through the above non-linear optimization, are highly sensitive to the initial conditions. In particular, when initialized with different assumptions, SQP can converge to different suboptimal local minima, resulting in discontinuities or abrupt variations in the motion of the assist branch over consecutive time steps.
To ensure motion smoothness, we use a B-spline interpolation to refine the trajectory. 

Given a set of \( n+1 \) control points \( \bm{P}_0, \bm{P}_1, \ldots, \bm{P}_n \), with \( \bm{P}_i \) being the joint state at time \( t_i \), and the degree of the spline \( p \), the B-spline trajectory \( \bm{C}(u) \) is defined as a weighted sum of basis functions. The parameter \( u \in [u_0, u_n] \) spans the curve domain, and the trajectory is expressed as
\begin{equation}
\bm{C}(u) = \sum_{i=0}^{n} \bm{P}_i \, N_{i,p}(u),
\end{equation}
where \( N_{i,p}(u) \) denotes the degree basis function \( p \) associated with the control point \( \bm{P}_i \)~\citep{unser1993b}.

The control points used in B-spline interpolation correspond to the discrete joint state obtained from Eq.~\eqref{Eq:SQP}. However, the resultant B-spline does not necessarily pass through these joint states. Instead, it prioritizes trajectory smoothness. As a consequence, the trajectory of the assist branch may no longer be the exact optimal solutions that minimize joint effort. However, as Eq.~\eqref{Eq:SQP} admits multiple feasible solutions, the smoothed states may still satisfy the dynamic constraints and remain within the feasible space.
To verify this, we introduce an additional optimization objective in the Sec.~\ref {sec:dyn_model} and Sec.~\ref{optimalcames} to ensure the feasibility of each candidate morphology. 


\section{Design Optimization: Morphology and Mounted Pose Determination}~\label{design_optimization}


This section presents the design approach that determines the optimal morphology and mounted pose with {the motion planner in the loop}. In particular, we consider both single-branch and bi-branch cases.

\subsection{Morphology Representation}~\label{morphology_representation}

\begin{figure*}[h]
  \centering
  \includegraphics[width=0.99\textwidth]{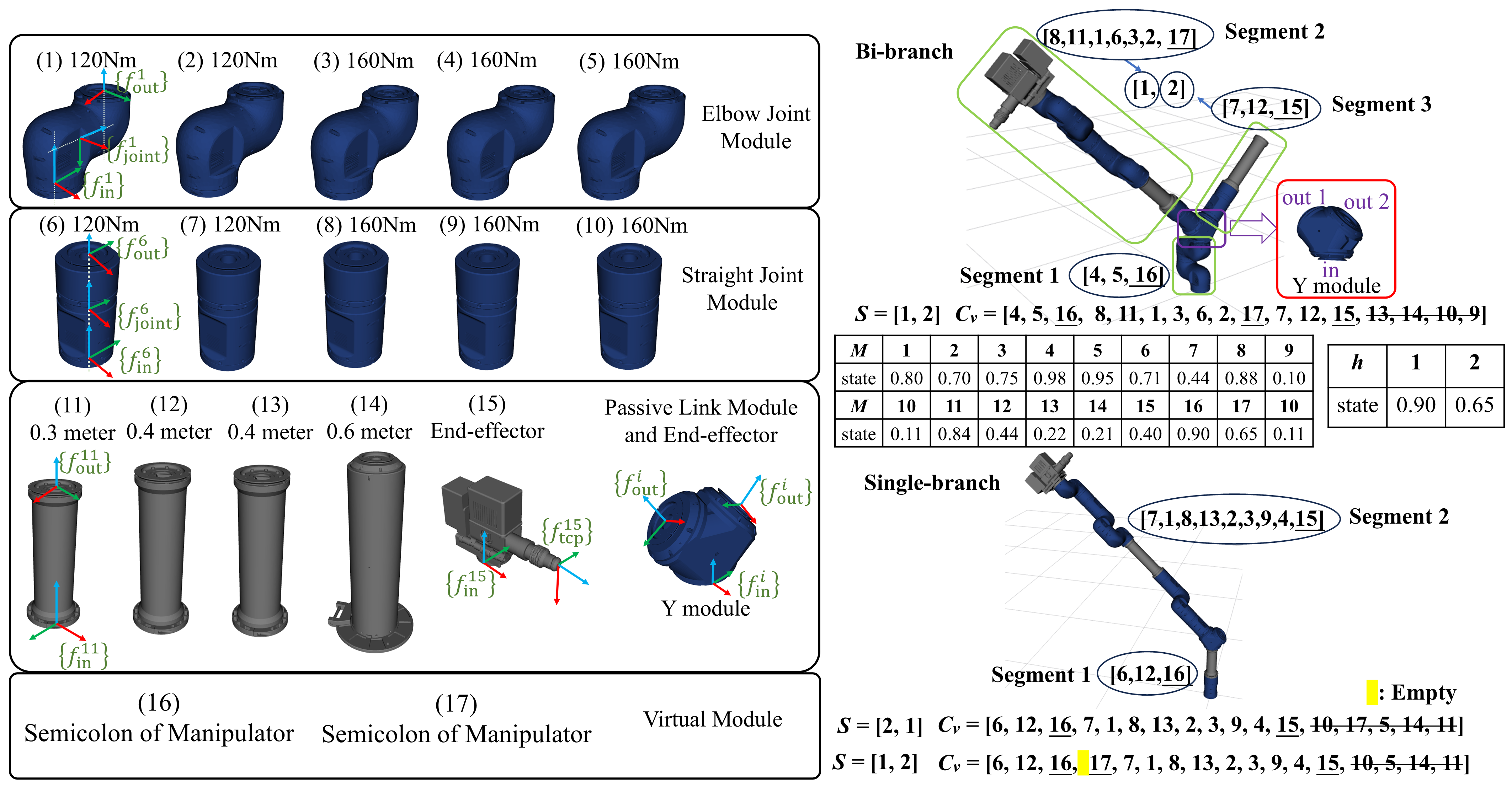}\vspace{-4mm}
  \caption{(Left) The available modules. 
  (Upper right) An example of a bi-branch morphology with its morphology state and modular state representation.  (Bottom right) An example of a single-branch morphology with its morphology state representation. 
  }
  \label{fig:representation}
\end{figure*}

 Each basic module of the modular manipulator is assigned a unique identifier (from the set \( \{1, 2, \dots, m\} \)) and the end effector is labelled as \( m+1 \). To realize multi-branch encoding, we introduce two virtual modules (named as semicolon of manipulator (SoM)) with IDs \( m+2 \) and \( m+3 \). The complete morphology is then encoded in a state vector \( \bm{C}_v \), which compactly represents module selection and assembly sequence, obeying the following rules:
\begin{enumerate}
    \item \textbf{Assembly encoding:}
    Each element in the vector \( \bm{C}_v \) marks one basic module, and the index defines the assembly sequence.
    \item \textbf{End of manipulator (EoM):}
    The end effector (marked by \( m+1 \)) terminates the kinematic chain. Modules listed after the EoM in \( \bm{C}_v \) are excluded from the final assembly.
    \item \textbf{Segment partition via SoM:}
    Virtual SoM modules (with ID \( m+2 \) and \( m+3 \)) partition \( \bm{C}_v \) into multiple segments. The first segment is mounted at the base, while the other segments work as operational branches. Note that when SoM modules appear, the 'Y' module (see Fig.~\ref{fig:representation}) that supports multiple branches is used and ID \( m+2 \) and \( m+3 \) mark the two output connectors of this `Y' module.

    
    \item \textbf{Mounted location via auxiliary vector:}
    An auxiliary vector \( \bm{S} \in \mathbb{Z}^2 \) specifies the mounting positions for the secondary branches. Each entry corresponds to a mounting port on the 'Y' module. The second segment is mounted at the first location marked by \( \bm{S} \), and the third segment (if any) is located at the second location.
\end{enumerate}

In this work, we assume \( 14 \) available modules (see Fig.~\ref{fig:representation}) with the end-effector assigned ID \(15 \). Two virtual SoM modules, assigned \(16 \) and \(17 \), are used to divide the morphology into multiple segments. 
Consider the example:
\(
\bm{C}_v \! = \! [4, 5, \textbf{16}, 8, 11, 1, 3, 6, 2, \textbf{17}, 7, 12, \textbf{\textit{15}}, \cancel{13, 14, 10, 9}]
\). Module 15 (the end-effector) appears at position 13, so all subsequent entries are excluded. The two SoMs (16 and 17) divide the vector into three segments, namely, segment 1 \( [4, 5]  \), segment 2 \([8, 11, 1, 3, 6, 2]\), and segment 3 \([7, 12]\). According to the representation rules, the first segment is always attached to the base. The appearance of SoMs triggers the insertion of a Y-connector, which provides two output ports. Given the auxiliary vector \( \bm{S} = [1, 2] \), the second and third segments are mounted on output ports 1 and 2, respectively. 
As shown in Fig.~\ref{fig:representation} (top right), three segments are partitioned by the Y-module: segment~1 is attached to the base, segment~2 to output port~1, and segment~3 to output port~2. Segment~2 always serves as the main branch of the bi-branch manipulator, with its end-effector equipped with a tool.

This encoding method allows for seamless switching between single- and bi-branch configurations. If the final vector includes only one or two non-empty segments, the morphology is interpreted as a single branch. 
For instance, the following two vectors:
\(
\bm{C}_v = [6, 12, \textbf{16}, 7, 1, 8, 13, 2, 3, 9, 4, \textbf{\textit{15}}, \cancel{10, 17, 5, 14, 11}]
\) and
\(
\bm{C}_v = [6, 12, \textbf{16}, \textbf{17}, 7, 1, 8, 13, 2, 3, 9, 4, \textbf{\textit{15}}, \cancel{10, 5, 14, 11}]
\),  
with mounting vectors \( \bm{S} = [2, 1] \) and \( \bm{S} = [1, 2] \), respectively, both correspond to the same single-branch manipulator (as shown in the bottom-right panel of Fig.~\ref{fig:representation}). In the first case, only one SoM (16) appears and the second segment is mounted through port 2 of the Y-connector. In the second case, two SoMs are present, but one segment between them is empty, resulting in a single-branch configuration. 


\subsection{Sorting-based Mapping Function} 
\label{map_function}

To enable continuous optimization, we introduce a map function \( g(\cdot) \) that transforms the discrete morphology state into a continuous variable. To this end, each module, including the end-effector, is assigned a continuous score in
\(
\bm{M} \in \mathbb{R}^{m+3},
\)
with \( M_i \in [0,1]\)\footnote{$M_i$ is generated by CMA-ES, as detailed in Sec.~\ref{optimalcames}.} .

Given the score vector \( \bm{M}\), the sorting map \( g(\bm{M}) \) ranks its components in descending order to generate the state vector:
\(
\bm{C}_v = g(\bm{M})
\).
That is, the module with the highest value (in $\bm{M}$) appears first, and the lowest value appears last. In this way, we implicitly encode both module selection and assembly order within a continuous optimization process. 
An example of this process is shown in the middle of Fig.~\ref{fig:representation}. Following the scenario in Sec.~\ref{morphology_representation}, consider the example with continuous state vector for modules \(1\) through \(17\):
\(
\bm M \) = [0.80, 0.70, 0.75, 0.98, 0.95, 0.71, 0.44, 0.88, 0.10, 0.11, 0.84, 0.44, 0.22, 0.21, 0.40, 0.90, 0.65
]. Sorting \(\bm M\) in descending order yields the index sequence with \(g(\bm M)\) =
\(
\bm {C}_v = [4, 5, \textbf{16}, 8, 11, 1, 3, 6, 2, \textbf{17}, 7, 12, \textbf{\textit{15}},\cancel{ 13, 14, 10, 9}]
\). 

The selection of mounting holes on the 'Y' module is also encoded using a continuous state vector \( \bm{h} \in \mathbb{R}^2 \). To this end, another sorting-based mapping function is applied to obtain the mounting configuration: \(
\bm{S} = g(\bm{h}),
\)
For example, given a continuous state \( \bm{h} = [0.90,\ 0.65] \), sorting in descending order yields
\(
g(\bm{h}) = \bm{S} = [1,\ 2],
\)
indicating that the second and third morphology segments are attached to output ports 1 and 2 of the 'Y' connector, respectively.



\subsection {Design Optimization Formulation}
The objective is to jointly optimize the morphology state \( \bm{C}_v \), the mounting hole order \( \bm{S} \), and the mounted pose \( \bm{P}_m \) of the modular manipulator to accomplish the desired tasks. Here, 
\(
\bm{P}_m = [x,\, y,\, z,\, \phi,\, \theta,\, \psi]^\top \in \mathbb{R}^6
\)
represents the six-dimensional pose of the manipulator base in the world frame \(\{W\}\), where \((x, y, z)\) denotes the Cartesian position and \((\phi, \theta, \psi)\) represent the roll, pitch, and yaw angles, respectively. With the given \(\bm{C}_V\),  \(\bm{S}\),  \(\bm{P}_m\), the joint states \( \bm{q}_i,\, \bm{\dot{q}}_i,\, \bm{\ddot{q}}_i \) in the \(i\)-th time step are computed by the HMPC.

\begin{figure}
  \centering
  \includegraphics[width=0.5\textwidth, trim=0pt 0pt 0pt 0pt, clip]{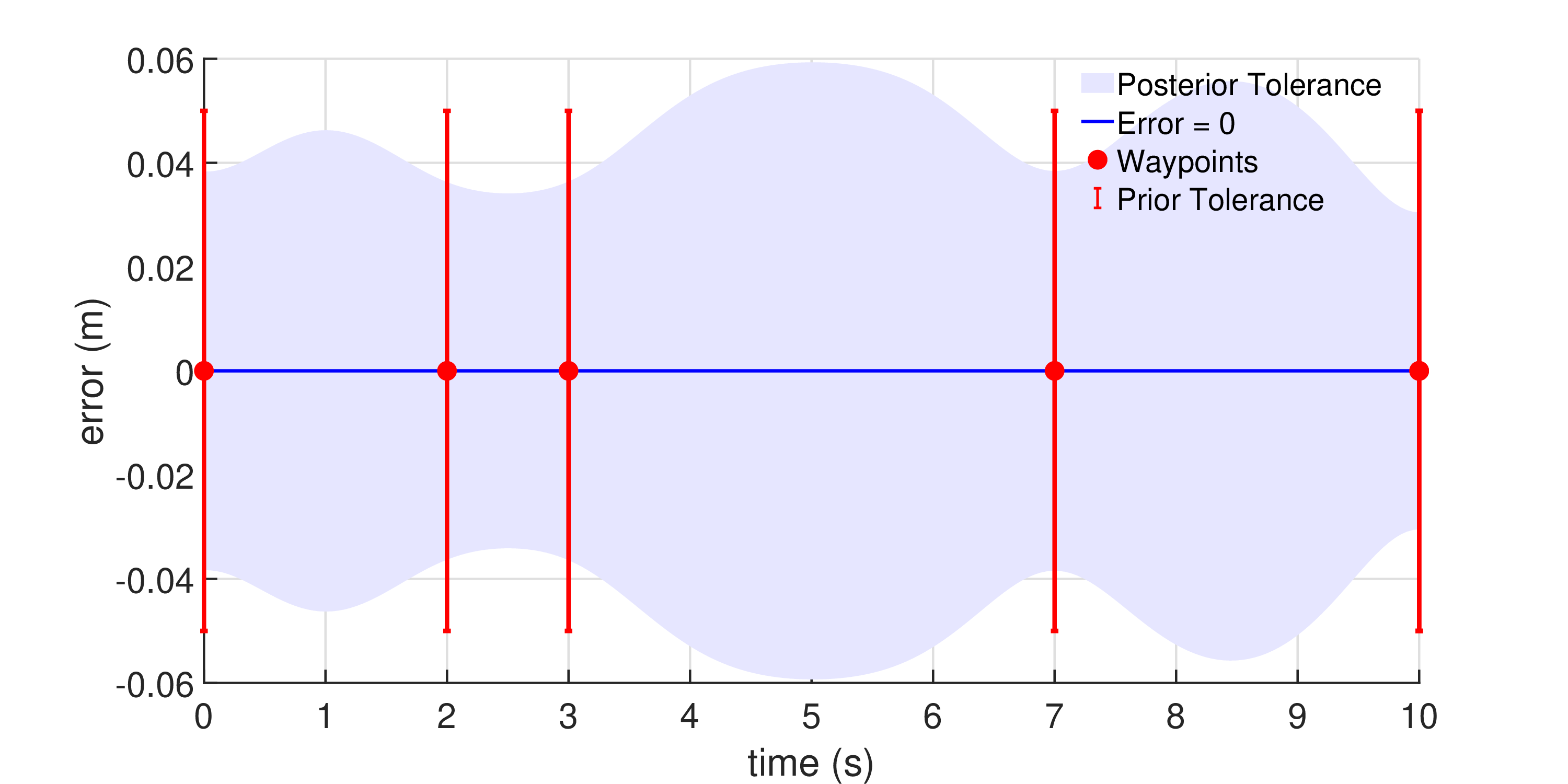}\vspace{-2mm}
\caption{\textbf{Tracking-error tolerance profile}. The prior value is \( \xi / 2 = 0.025 \) at the specific waypoint. The blue solid line indicates the posterior mean (set to 0), and the shaded region represents the 95\% confidence interval of the posterior tolerance across all time steps.}
  \label{fig:GP_process_tracking}
\end{figure}

\subsubsection{Task execution requirement and objective}
\ \

\textit{Requirement:}
The reference trajectory \(\bm{P}(t)\) and \(\mathbf{R}(t)\) are generated by using pre-defined key way-points. 
In real-world scenarios, precise tracking is typically required only at specific key waypoints. At these waypoints, the tracking error bounds are defined for position and orientation. The required tracking error at a specific time \(t\), corresponding to a desired key position and orientation, can be expressed as
\begin{equation}
\begin{aligned}
- \bm{\xi}_p(t) &\le \bm{p}_t - \bm{p}_{d,t} \le \bm{\xi}_p(t), \\
- \bm{\xi}_o(t) &\le \log\left( \mathbf{R}_t \mathbf{R}_{d,t}^\top \right)^\vee \le \bm{\xi}_o(t),
\end{aligned}
\label{eq:tracking_error_split_bounds}
\end{equation}
where \(\boldsymbol{\xi}(t) = \begin{bmatrix} \boldsymbol{\xi}_{p}^{T}(t) \ \ \boldsymbol{\xi}_{o}^{T}(t) \end{bmatrix}^{T} \in \mathbb{R}^{6}\) defines the tracking tolerance vector at time \(t\).

Given reference trajectory \(\bm{P}(t)\) and \(\mathbf{R}(t)\), and the tracking error bounds defined at specific waypoints, we use Gaussian process regression (GPR) to generate smooth, time-varying bounds over the entire trajectory. In particular, we set zero error (with high confidence) at each anchor point and use the tracking tolerance as the prior variance. The resultant 95\% confidence interval yields a smooth, adaptive error bound—tight near the anchor waypoints and more relaxed elsewhere\footnote{For further details, please refer to 
Appendix~\hyperref[sec:gp_fitness]{A} }
{, Fig.~\ref{fig:GP_process_tracking} presents a representative
result of the GPR-based error bounds. The shaded blue area depicts the 95\% predictive interval \( \bm{\xi} = 2\sigma_i(t) \)~\citep{casella2024statistical, williams2006gaussian}. The solid red line indicates the confidence interval 95\% of the prior distribution, while the blue region represents the tolerance limits varying over time. As discussed in~\citep{williams2006gaussian}, the posterior variance at the training points is strictly lower than the prior variance, ensuring that the tolerance limits at the waypoints remain within the predefined bounds.}

Then, at each time step, the interpolated constraints on tracking errors can be expressed as
\begin{equation}
\begin{aligned}
-\bm{\bar{\xi}}_p(t_i) \le \bm{p}_i - \bm{p}_{d,i} \le \bm{\bar{\xi}}_p(t_i), \\
-\bm{\bar{\xi}}_o(t_i) \le \log\left( \mathbf{R_i} \mathbf{R_{d,i}}^\top \right)^\vee \le \bm{\bar{\xi}}_o(t_i),
\end{aligned}
\label{eq:tracking_error_split_bounds_gp}
\end{equation}
where \( \boldsymbol{\xi}(t) = [\boldsymbol{\bar{\xi}}_{p}^{\top}(t) \ \ \boldsymbol{\bar{\xi}}_{o}^{\top}(t)]^{\top}= 2\boldsymbol{\sigma}(t) \in \mathbb{R}^{6} \) represents the time-varying error bounds derived from the posterior standard deviation \( \boldsymbol{\sigma}(t) \) of the GPR model.

\begin{figure}
  \centering
  \includegraphics[width=0.5\textwidth, trim=0pt 0pt 0pt 0pt, clip]{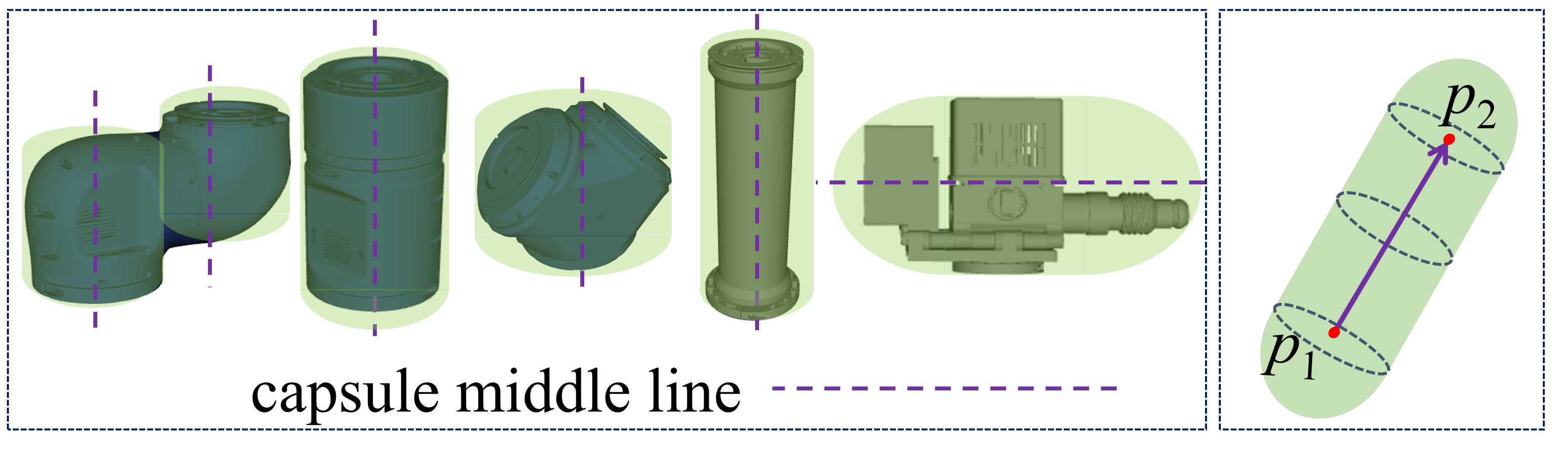}\vspace{-2mm}
\caption{(\text{Left}) Capsule model used for collision a for the different modules. (\text{Right}) Mathematical representation of a capsule formed by sweeping a sphere from point \(p_1\) to point \(p_2\).}
  \label{fig:capsule}
\end{figure}

\textit{Objective:}
We define the desired end-effector pose at the \(i\)-th waypoint as \(\mathbf{T}_{d,i} \in \mathrm{SE}(3)\). The task execution performance is evaluated by averaging the tracking error over all time steps, which can be expressed as
\begin{equation}
\frac{1}{N} \sum_{i=1}^{N} 
\left\| 
\mathrm{FK}(\bm{C}_v, \bm{P}_m, \bm{S}, \bm{q}_i)\, 
\mathbf{T}_{d,i}^{-1} 
\right\|^2,
\label{eq:tracking_error}
\end{equation}
where \(\mathrm{FK}(\cdot)\) denotes the forward kinematics, \(\bm{q}_i\) is the joint state at time step \(i\), and \(N\) is the total number of time steps.



%
\subsubsection{Dynamic constraints}~\label{sec:dyn_model}
In the HMPC formulation, dynamic constraints such as actuator torque limits are omitted, compromising the feasibility. In this stage, we integrate dynamic constraints into the optimization process to address this limitation. Given the planned joint position \( \bm q_i \), velocity \( \dot{\bm q}_i \), acceleration \( \ddot{\bm q}_i \), and end-effector force \( \bm F_{\mathrm{ext},i} \) in the \( i \)-th step, the required joint torques are computed as
\begin{equation}
\bm \tau_i 
\!=\! \mathbf{M}(\bm q_i)\,\ddot{\bm q}_i
\!+\! \mathbf C(\bm q_i,\dot{\bm q}_i)\,\dot{\bm q}_i
\!+\! \bm G(\bm q_i)
\!-\! \mathbf J(\bm q_i)^\top \bm F_{\mathrm{ext},i},
\label{eq:inverse_dynamics}
\end{equation}
where \( \mathbf{M}(\bm q_i) \) is the inertia matrix, \( \mathbf C(\bm q_i, \dot{\bm q}_i) \) is the Coriolis and centrifugal term, \( \bm G(\bm q_i) \) denotes gravity, and \( \mathbf J(\bm q_i) \) is the Jacobian at the end-effector.

%
To ensure dynamic feasibility, the computed joint torques must satisfy the actuation limit. For the \( j \)-th joint, we have
\begin{equation}
\bigl| \tau_{i,j} \bigr| \;\le\; \tau_{\max,j}, \quad j = 1, \dots, n_j,
\label{eq:torque_constraints}
\end{equation}
where \( \tau_{i,j} \) denotes the \( j \)-th element of the torque vector \( \bm \tau_i \), and \( \tau_{\max,j} \) is the maximal allowable torque for joint \( j \) that is defined by the hardware.  The total joint number \(n_j\) of the manipulator with the given morphology
will be further optimized in Sec.~\ref{optimalcames}. 


\subsubsection{Collision avoidance}
In this stage, collision checking, including self-collision and collision with the environment, is integrated into our optimal design process to realize collision avoidance.

To enable efficient collision detection, each module is approximated as a \textit{capsule}—a line segment with a uniform radius, providing a coarse yet reasonable geometric abstraction (see Fig.~\ref{fig:capsule}). Each capsule is parameterized by endpoints \(\bm{p}_1\), \(\bm{p}_2\), and radius \(r\), with central axis, defined by
\begin{equation}
\bm{P}(t) = \bm{p}_1 + t(\bm{p}_2 - \bm{p}_1), \quad t \in [0, 1].
\label{eq:capsule_axis}
\end{equation}

\textit{Collision avoidance with the environment}: 
We compute a signed distance field (SDF) of the workspace~\citep{oleynikova2016signed} to detect possible collisions, enabling fast distance and gradient queries. Given the joint state \(\bm q_i\), the pose of each capsule is computed in the world frame. Sampling points \(\bm p_r\) along each capsule middle line, we query the SDF and enforce: 
\begin{equation}
\mathrm{SDF}\bigl(\bm p_r(\bm q_i)\bigr)
\;\ge\; r + d_{\mathrm{safe}},
\quad
\forall\,\bm p_r \in \mathcal{C}_t(\bm q_i),
\end{equation}
where \(\mathcal{C}_t(\bm q_i)\) is the set of sampled points on the \(t\)-th module capsule, and \(d_{\mathrm{safe}}\) is a safety margin. 


{\textit{Self-collision avoidance:} }
Consider two such capsules defined by segments \([\bm{p}_1^{m}, \bm{p}_2^{m}]\), \([\bm{q}_1^{m}, \bm{q}_2^{m}]\) and radii (\(r_p\), \(r_q\)), the minimal distance between them is
\begin{align}
d_{\min} &= \min_{t_p,\,t_q \in [0,1]} \;\; \left\| \bm{p}^{m}(t_p) - \bm{q}^{m}(t_q) \right\| \label{eq:capsule_dist} \\
\text{s.t.} \quad
\bm{p}^{m}(t_p) & = \bm{p}_1^{m} + t_p (\bm{p}_2^{m} - \bm{p}_1^{m}), \nonumber \\
\bm{q}^{m}(t_q) & = \bm{q}_1^{m} + t_q (\bm{q}_2^{m} - \bm{q}_1^{m}). \nonumber
\end{align}

Then, the self-collision avoidance is realized when the following constraint is satisfied:
\begin{equation}
d_{\min} > r_p + r_q + d_{\mathrm{safe}},
\end{equation}
where \(d_{\mathrm{safe}}\) is a safety margin that ensures sufficient clearance.
For general self-collision checking between all robot modules, we impose:
\begin{equation}
\! \! \!\! \!  C\bigl(\bm{p}_r^{m}, \bm{q}_i^{m}\bigr) \geq d_{\mathrm{safe}}, \!\!\!\!\! \quad \forall\, \bm{p}_r \! \in \!  \text{Robot surface},\; \! i = 1,\dots,N
\end{equation}
where \(C(\bm{p}^{m}_r, \bm{q}^{m}_i)\) denotes the minimum distance \(d_{\text{min}}\) between any pair of modules at the joint state \(\bm{q}_i\).


\subsubsection{Functional metrics}

We further introduce two functional metrics to enhance the performance: \textit{1)} minimizing the joint effort, and \textit{2)} maximizing the manipulability.

After computing the desired joint torques \(\tau_{i,j}\) via Eq.~\eqref{eq:inverse_dynamics}, we then minimize the cost
\begin{equation}
F_{\mathrm{eff}}
= \frac{1}{N}\sum_{i=1}^{N}\sum_{j=1}^{n_j}\tau_{i,j}^{2} ,
\end{equation}
where 
\(N\) is the number of time steps.

The second metric aims to maximize {manipulability}, which is defined as
\begin{equation}
    \ M_{\text{man}} = \frac{1}{N} \sum_{i=1}^{N} \det\left(\mathbf{J}(\bm{q}_i)\, \mathbf{J}^\top(\bm{q}_i)\right),
\end{equation}
where \( \mathbf{J}(\bm{q}_i) \) is the Jacobian matrix at configuration \( \bm{q}_i \).

When the number of joints equals or is smaller than the task's dimensions, the Jacobian matrix \( \mathbf{J}(\bm{q}_i) \in \mathbb{R}^{6 \times d_m} \) lacks redundancy. As a result, the Jacobian may become rank-deficient, causing the manipulability value \( \det\left(\mathbf{J}(\bm{q}_i)\, \mathbf{J}^\top(\bm{q}_i)\right) \) to vanish. To overcome this limitation, the task-specific sub-Jacobian \( \mathbf{J}_{\text{sub}}(\bm{q}_i) \in \mathbb{R}^{d_h \times d_m} \) is extracted for calculating the manipulability, where \( d_h \) is the number of dimensions associated with the high-priority task.
The manipulability for this sub-task is then defined as
\begin{equation}
    M_{\text{man, sub}} = \frac{1}{N} \sum_{i=1}^{N} \det\left(\mathbf{J}_{\text{sub}}(\bm{q}_i)\, \mathbf{J}^\top_{\text{sub}}(\bm{q}_i)\right).
\end{equation}

Note that, for bi-branch morphologies, the manipulability is only computed for the main branch, while the joint effort reduction is enforced on all branches.


\subsection{CES-MS-based Optimization}~\label{optimalcames}
The primary objective of the design optimization is to ensure successful task execution while satisfying all constraints. To this end, we solve the following problem,
\begin{equation}
\begin{aligned}
    & \underset{\bm{C}_v,\bm{P}_m, \bm{S}}{\text{min}}
    & & \! \!\! X_e = -w\, e^{-w_f F_{\text{eff}} + w_m M_{\text{man}}} + w_l\, \mathrm{len}(\bm{C}_v) \\
    & \text{subject to}
    & &  \text{Redundancy}\;\left\{
\begin{aligned}
        \operatorname{dim}(\mathcal{T}_H) &< \operatorname{dim}(\mathcal{R}),\\
        \operatorname{dim}(\mathcal{T}_L) &< \operatorname{dim}(\mathcal{R}),
\end{aligned}
\right. \\
    & & & -\bm{\bar{\xi}}_p(t) \le \bm{p}_i - \bm{p}_{d,i} \le \bm{\bar{\xi}}_p(t),  \forall i \\
    & & & -\bm{\bar{\xi}}_o(t) \le \log\left( \mathbf{R}_i \mathbf{R}_{d,i}^\top \right)^\vee \le \bm{\bar{\xi}}_o(t),  \forall i \\
    & & & C(\bm{p}_r, \bm{q}_i) \ge d_{\text{safe}}, \quad \forall \bm{p}_r \in \text{Robot},\, \! \forall i \\
    & & & \mathrm{SDF}\bigl(\bm p_r(\bm q_i)\bigr) \! \!\! 
\;\ge\; \! \!\!  r + d_{\mathrm{safe}},
\quad  \!\!\!\!\!\! 
\forall\,\bm p_r \in \mathcal{C}_t(\bm q_i),   \forall i \\
    & & & \|\bm{\tau}_i(\bm{q}_i, \dot{\bm{q}}_i, \ddot{\bm{q}}_i)\| \le \bm{\tau}_{\text{max}},  \forall i \\
    & & & F_{\text{eff}} = \frac{1}{N} \sum_{i=1}^{N} \sum_{j=1}^{d} \tau_{i,j}^{2}, \\
    & & & M_{\text{man}} = \frac{1}{N} \sum_{i=1}^{N} \det\left( \mathbf{J}_{\text{sub}}(\bm{q}_i) \mathbf{J}_{\text{sub}}^\top(\bm{q}_i) \right),
\end{aligned}
\label{eq:constraint}
\end{equation}
where \( w \), \( w_f \), \( w_m \), and \( w_l \) are weighting factors. The term \( \mathrm{len}(\bm{C}_v) \) penalizes the module usage. 

In the above formulation, `Redundancy' constraints ensure that the number of DoFs exceeds the task dimensionality for both high- and low-priority sub-tasks.



The original morphology optimization involves selecting and arranging modules from a discrete space, choosing a mounting hole, and adjusting the mounted pose in a continuous space, which naturally results in a mixed-integer nonlinear programming (MINLP)~\citep{sahinidis2019mixed} that is challenging to solve. In this work, instead of directly optimizing the discrete morphology \( \bm{C}_v \) and mounted position state \( \bm{S} \), we optimize continuous vectors \( \bm{M} \) and \( \bm{h} \), as defined in Sec.~\ref{map_function}. This mapping reformulates the original optimization problem into a fully continuous optimization problem, enabling joint optimization of morphology and mounted pose within a general optimization framework\footnote{Readers are suggest to to check the simple example provided in Appendix~\hyperref[cmaes_anylsis_MINLP]{B} for more details.}.
%
%
As a result, the optimization problem can be defined as
\begin{equation}
\begin{aligned}
\! \! \underset{g(\bm{M}),\, \bm{P}_m,\, g(\bm{h})}{\text{min}}  
& \!\!\! \!\! \!\!\! E \! = \! E_{\text{track}} \!+\! E_{\text{col}} \!+ \!E_{\text{dyn}} \!+\! E_{\text{red}} \!+\! w_l\,\mathrm{len}(\bm{C}_v) \\
&\quad - \delta \cdot w \exp(-w_f F_{\text{eff}} + w_m M_{\text{man}})
\end{aligned}
\label{eq:cmaesopt}
\end{equation}
where, 
\begin{equation}
\delta =
\begin{cases}
1, \! \! \!\!\! & \text{if Eq.~\eqref{eq:tracking_error_split_bounds_gp} holds and } E_{\text{col}} \! = \! E_{\text{dyn}} = \! E_{\text{red}} \!= \! E_{\text{tra}} \! = \! 0 \\
0, \! \! \!\!\! & \text{otherwise}
\end{cases}
\label{eq:delta_condition}
\end{equation}

The cost terms are defined as
\begin{equation}
\begin{aligned}
E_{\text{track}}
& = \frac{1}{N} \sum_{i=1}^{N} 
\left\| 
\mathrm{FK}(\bm{C}_v, \bm{P}_m, \bm{S}, \bm{q}_i)\, 
\bm{\mathbf{T}}_{d,i}^{-1} 
\right\|^2,
\\
E_{\text{tra}} &= 
\begin{cases}
0, & \text{error below threshold at each waypoint}, \\
\infty, & \text{otherwise}
\end{cases}
\\
E_{\text{col}} &= 
\begin{cases}
\infty, & \text{if any collision is detected}, \\
0, & \text{otherwise}
\end{cases}\\
E_{\text{dyn}} &= 
\begin{cases}
\infty, & \text{if any } \|\bm{\tau}_i\| > \bm{\tau}_{\max}, \\
0, & \text{otherwise}
\end{cases}\\
E_{\text{red}} &= 
\begin{cases}
0, & \text{if redundancy is satisfied}, \\
\infty, & \text{otherwise}
\end{cases}
\end{aligned}
~\label{eq:cames_constrants}
\end{equation}
where \( E_{\text{track}} \) quantifies tracking error across all time steps.

{The above optimization problem is then solved using the CMA-ES algorithm~\citep{krause2016cma} , which operates in a continuous domain. With the motion planner in the loop, candidate designs are evaluated via customized metrics, and the algorithm iteratively updates \(\bm{M}\), \(\bm{h}\), and \(\bm{P}_m\) toward optimal solutions. The final morphology is obtained through \( \bm{C}_v = g(\bm{M}) \) and \( \bm{S} = g(\bm{h}) \). At each generation, the fitness function (Eq.~\eqref{eq:cmaesopt}) is evaluated by incorporating multiple criteria (see Eq.~\eqref{eq:cames_constrants}), ensuring that the optimal design satisfies tracking accuracy, collision avoidance, and dynamic constraints before being optimized for minimal joint effort and maximal manipulability, as indicated by Eq.~\eqref{eq:delta_condition}.
}

It is important to note that, to distinguish identical modules during optimization, each basic module is assigned a unique identifier—even if they are physically of the same type. {For example, as illustrated in Fig.~\ref{fig:representation}, modules 1–3, 4–5, 6–8, and 9–10 are identical, but are assigned with different identification number.} Therefore, we do not need to explicitly impose the constraint on the number of available modules in the optimization formulation.

\section{Evaluation}~\label{experimentandsimulation}

This section validates the proposed framework by conducting extensive simulations and hardware experiments. All the results can be addressed at \url{https://youtu.be/2KI7wOQjXAo}.


\subsection{Implementation Details}

\paragraph{HMPC and SQP parameters.}
For both high-level and low-level HMPC, the tracking gains $\mathbf{Q}_k$ (in Eq.~\eqref{eq:highlevel_MPC} and Eq.~\eqref{eq:low_level_MPC}) are set to 10 for position and 4 for orientation. The regularization weights \(\mathbf{R}_k\) for all variables are set to 0.005, with the time step \( dt = 0.01 \) s. The prediction horizons are set to \(N_h = N_l = 10\), and each quadratic program is solved using the off-the-shelf solver \texttt{qpOASES}~\citep{ferreau2014qpoases}. In HMPC, we start by solving IK to obtain a joint state corresponding to the initial task-space waypoint, serving as the starting state. 
For SQP-based refinement, the regularization and penalty parameters are set to \(\lambda = 0.01\) and \(\mu = 100\), respectively. Note that above setups are fixed across all tests.

{
\paragraph{CMA-ES setup: } 
The maximal population size is 40, and the initial sampling standard deviation ($\sigma$) is 0.25. 
Each variable is normalized in the range \([0, 1]\). 
The algorithm is initially set to run for 100 generations. 
If, after 100 generations and the condition $\sigma < 0.005$ is not satisfied, the evolution continues until convergence is reached for 200 generations. { For the objectives in Eq.~\eqref{eq:cmaesopt}, we have \( w = 1 \) and \( w_l = 0.001 \). The evaluation process for different populations at each generation is executed by using the off-the-shelf solver \texttt{pycma}~\citep{hansen2019pycma}. }
}


\paragraph{Hardware setup:}
The basic modules used in the current research are illustrated in Fig.~\ref{fig:representation}. The maximal joint torques are 120~Nm for Modules~1, 2, 4, and 5, and 160~Nm for Modules~3 and 6. For each module, the joint position, angular velocity and acceleration are separately constrained within \(\pm2.4\) $\mathrm{rad}$, \(\pm2.0\) \(\mathrm{rad/s}\) and \(\pm0.5\) \(\mathrm{rad/s}^2\). Compared with the previous work~\citep{lei2024task}, the total number of available modules (excluding end-effectors) increased from 9 to 15, enabling the construction of bi-branch morphologies. As a result, the number of feasible manipulator configurations expands significantly from approximately \(2.9 \times 10^5\) to \(6.2 \times 10^6\), resulting in a substantially larger design space.



\subsection{Pick-and-place with Obstacle Avoidance}

\subsubsection{Task description}
\begin{figure*}
  \centering
  \centering  \includegraphics[width=0.99\textwidth]{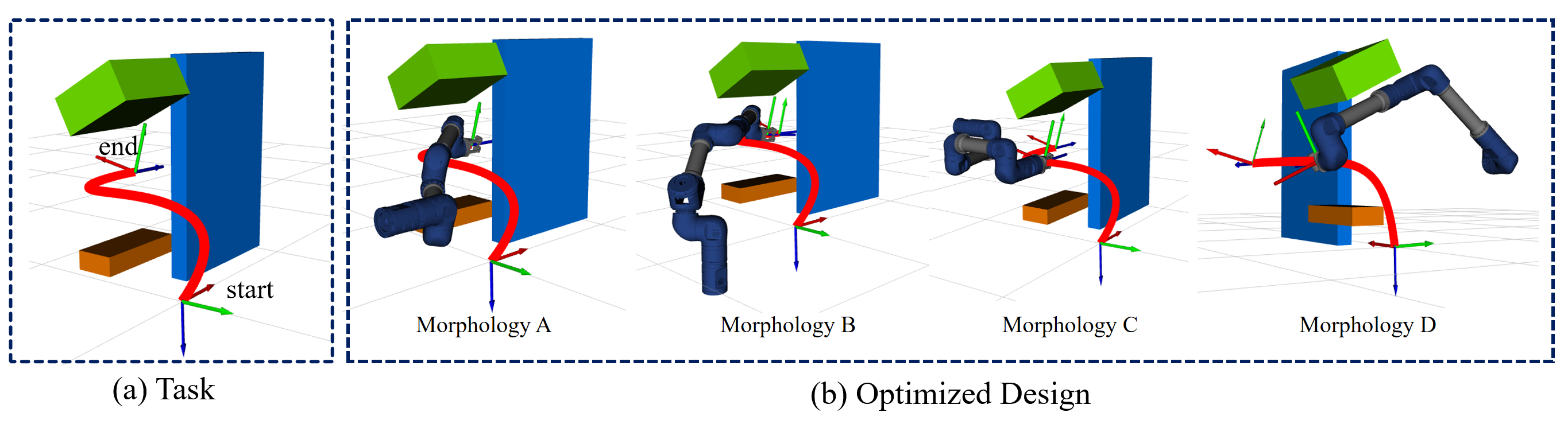}\vspace{-4mm}
  \caption{\textbf{Optimized designs and the movements for pick-and-place task}: (Left) Desired trajectory in the cluttered environment, where the blue obstacle is located between the start and goal point, (Right) Four optimal morphologies and the corresponding movements for the pick-and-place task.}
  \label{fig:simulation_env_pickandplace}
\end{figure*}

In the pick-and-place task, the manipulator grasps an object and moves in a cluttered environment containing three box-shaped obstacles (marked by blue, green, and yellow), as shown in Fig.~\ref{fig:simulation_env_pickandplace} (left). The object is moved from the start point \([0, 0, 0]\) $\mathrm{m}$ with orientation \([\pi,\, 0,\, 0]\) $\mathrm{rad}$ to the goal point \([0.5, 0.7, 0.5]\) $\mathrm{m}$ with orientation \([0,\, -\pi/15,\, \pi]\) $\mathrm{rad}$. 
In this task, position tracking is prioritized over orientation. To accomplish the desired task, a collision-free task-space trajectory is generated between the two waypoints. The allowable errors are set to \(0.0005\,\mathrm{m}\) for position and \(0.001\,\mathrm{rad}\) for orientation, corresponding to a Gaussian prior with zero mean and 95\% confidence intervals. 
For the optimization objectives in Eq.~\eqref{eq:cmaesopt}, joint effort minimization and manipulability maximization are both considered, with \( w_f = 0.01 \) and \( w_m = 5 \), respectively.


\begin{table*}[h]
\centering
\begin{tabular}{|c|c|c|c|c|c|c|c|}
\hline
\textbf{Guess} & 
\textbf{Morphology} & 
\textbf{Mounted Pose} & \textbf{Effort} & \textbf{Manipulability} & \textbf{Overall Cost} & \textbf{DoFs} \\ \hline
1 & A &[0.58, 0.19, 0.43, 1.57, 0.01 , 0.02] & 112.88 & 0.016 & -1.1128 & 4 \\ \hline
2 & B &[0.02, -0.48, 0.08, 0.0, -0.2 , 1.56] & 320.50 & 0.256 & -2.9490 & 5 \\ \hline
3 & C &[0.00, -0.79, 0.57, 0.0, -3.14 , 1.57] & 268.63 & 0.090 & -2.5963 & 5 \\ \hline
4 & D &[0.15, 1.38, 0.51, 0.0, -1.56 , 1.58] & 173.53 & 0.059 & -1.6763 & 5 \\ \hline
\end{tabular}
\caption{\textbf{Optimized results obtained with different initial guesses for the pick-and-place task.} The table shows evaluated results include effort, manipulability, optimized  DoFs number and the combined cost which was computed by \(-w_f F_{\text{eff}} + w_m M_{\text{man}}\) where \(w_f = 0.01\) and \(w_m = 5\).}
\label{tab:morphology_comparison_pickaandplace}
\end{table*}
\begin{figure*}
  \centering
\includegraphics[width=0.99\textwidth]{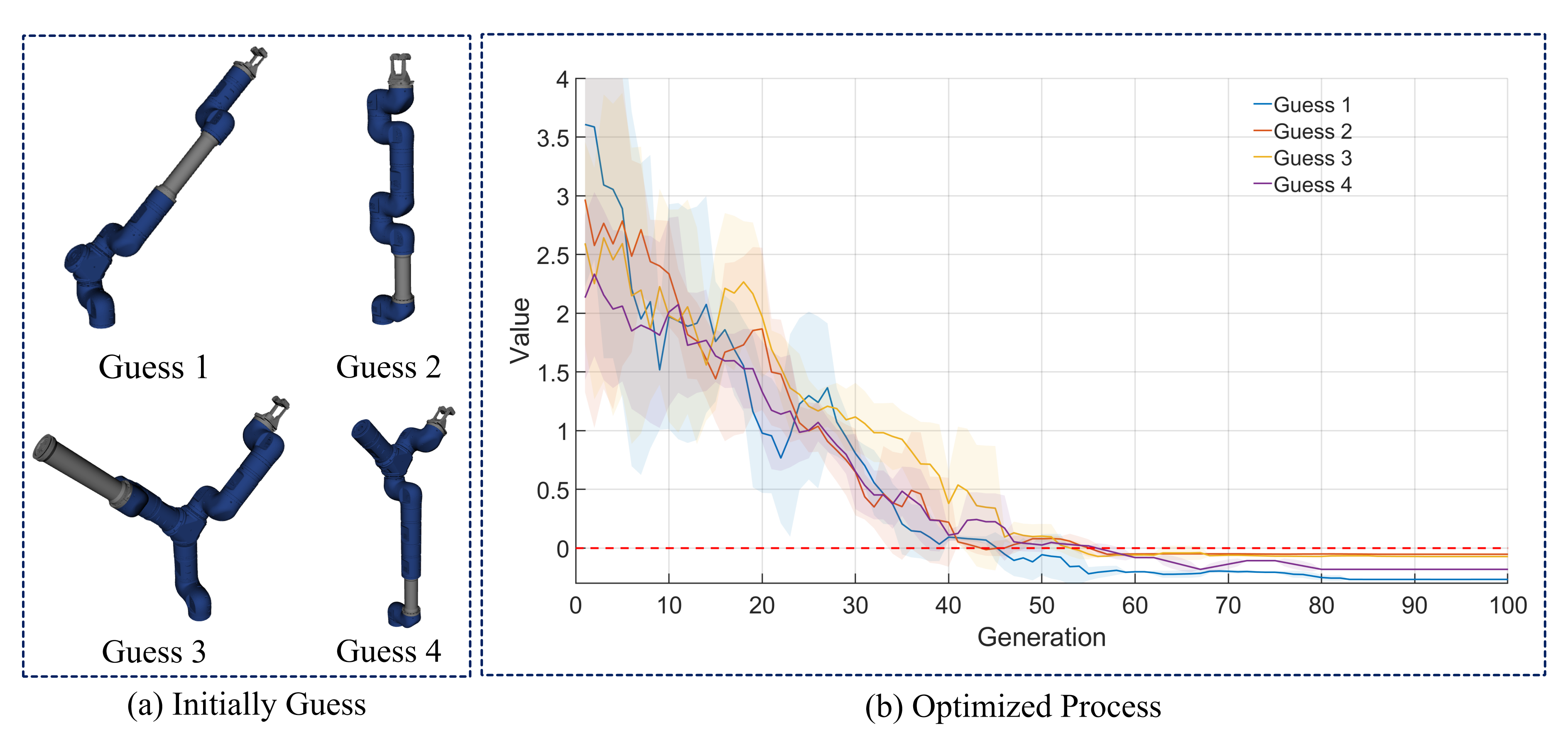}\vspace{-4mm}
  \caption{ \textbf{Optimal design process for executing pick-and-place task: }(Left) Optimized morphologies with different initial guesses; 
  (Right) The four curves represent distinct optimization processes, each initiated with a different initial guess. The solid lines depict the moving averages of the minimum evaluation values, and the shaded areas illustrate their corresponding rolling standard deviations.}
  \label{simulation}
\end{figure*}
\begin{figure*}
  \centering
  \centering  \includegraphics[width=1.0\textwidth, trim=50pt 0pt 50pt 0pt, clip]{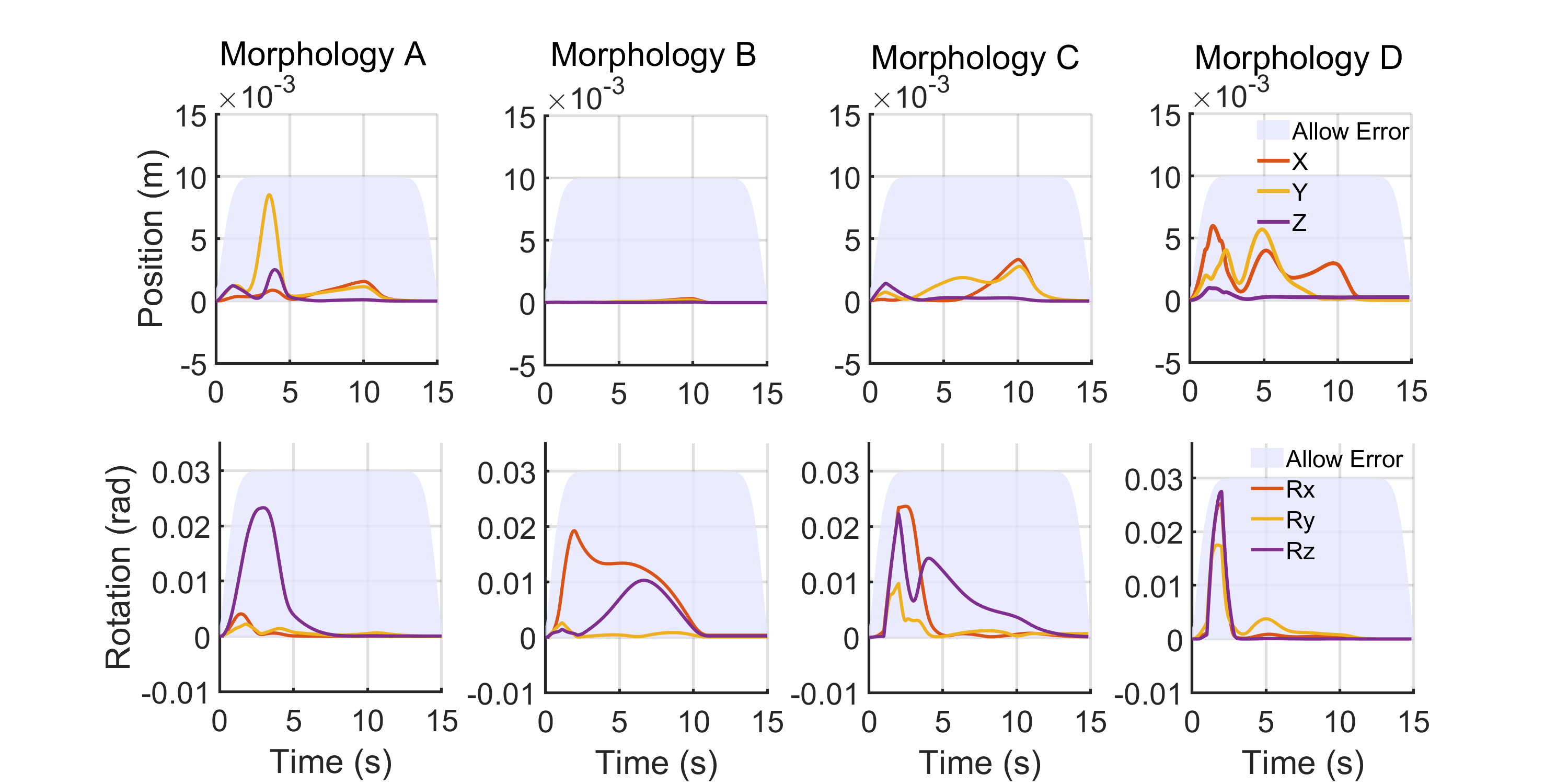}\vspace{-2mm}
  \caption{\textbf{Absolute tracking errors of the four optimal designs}. The blue shaded region represents the allowable tracking error boundary defined by the GPR-based interpolation. The first row illustrates the positional tracking errors along the \(x\), \(y\), and \(z\) axes, while the second row shows the orientation tracking errors. In this figure, only positive error magnitudes are shown due to the use of absolute values. }
  \label{fig:all_tracking_multi_object}
\end{figure*}

\subsubsection{Optimization with different initial guess}
We run the optimization with four distinct initial guesses to evaluate the robustness of the proposed computational design framework, where each guess includes a specific combination of modular state \( \bm{M} \), mounting hole state \( \bm{S} \), and mounted pose \( \bm{P}_m \). As illustrated in Fig.~\ref{simulation} (left), the guesses include two with dual branches and two with single branches. The evolution of the minimum cost across iterations is shown in Fig.~\ref{simulation} (right). It turns out that in all four cases, CMA-ES finds an optimal solution within 100 iterations. In particular, the negative cost (after 60 generations) indicates successful task execution, i.e., meeting the accuracy requirements and satisfying dynamic and collision constraints. The final results, including the optimized structures and poses, are visualized in Fig.~\ref{fig:simulation_env_pickandplace}, with Morphology~A to D, corresponding to the initial guess~1 to 4. Detailed optimal results are reported in Table~\ref{tab:morphology_comparison_pickaandplace}.


As shown in Table~\ref{tab:morphology_comparison_pickaandplace}, one of the generated manipulators has four DoFs, while the other three have five. 
Since the high-priority sub-task involves position tracking in three directions, all final designs meet the HMPC formulation requirement with DoFs greater than the minimum threshold of three.
Meanwhile, since the optimization formulation penalizes excessive module usage, the optimized morphologies remain efficient without being overly redundant. Regardless of the initial structure—whether single-branch or bi-branch morphology—the optimization consistently converges to a single-branch morphology. 

Furthermore, we evaluate the tracking performance of the different optimized solutions. In this task, we require the tracking errors at both the initial and final trajectory points to remain within predefined tolerance bounds. Fig.~\ref{fig:all_tracking_multi_object} visualizes the real errors, where the blue-shaded regions denote the allowable threshold obtained by GPR. It turns out that, due to the penalty on tracking errors ($E_{\text{track}}$ in Eq.~\eqref{eq:cmaesopt}) and the explicit constraints on waypoint tracking accuracy, all trajectory errors remain within the permitted bounds (shaded regions in Fig.~\ref{fig:all_tracking_multi_object}) throughout the planned motion. 
The simulation video demonstrating execution of the pick-and-place task with four morphologies in a cluttered environment with multiple obstacles is provided in the supplementary material. 

\subsection{Polishing Task} 
\subsubsection{Task description}
This section evaluates the efficacy of a car-door polishing task, a routine process in automotive manufacturing that aims to remove surface defects from metal panels. In particular, we test three different objective preferences by varying the weights in the objective function:
\begin{itemize}
    \item \textbf{Scenario 1}: Joint optimization of maximizing manipulability and minimizing joint effort, with weights \( w_m = 5.00 \), \( w_f = 0.01 \).
    \item \textbf{Scenario 2}: Focus solely on maximizing manipulability, with \( w_m = 5.00 \), \( w_f = 0.00 \).
    \item \textbf{Scenario 3}: Focus solely on minimizing joint effort, with \( w_m = 0.00 \), \( w_f = 0.01 \).
\end{itemize}

{
In this task, the objective is to follow a predefined trajectory defined in Cartesian space that has the $z$-axis normal to the surface. Furthermore, since the tool is symmetric, rotation about the local \( z \)-axis is negligible. 
To generate a reference trajectory obeying the requirement, we rely on the nominal shape of the workpiece, described as a polyhedral mesh, which 
can be easily retrieved from CAD models.
To generate a dense boustrophedon path, i.e., a serpentine-like path, on the mesh, we rely on geodesic line interpolation~\citep{chenier1996shortest}.
Given the two extremal points of each motion segment, we construct the on-surface shortest path connecting those points to build a dense trajectory of points on the surface. Exploiting the geometry information, we fixed the orientation with the $z$-axis normal to the surface, while $x,y$-axes are optimized on the whole trajectory to minimize the rotation. 
As full-coverage motion planning exceeds the scope of this work, we manually adjusted the endpoints for the different segments of the serpentine-like path. The generated polished positional trajectory on the mesh is shown in Fig.~\ref{fig:polish} (left). 
}

To assess tracking performance, GPR is used to model the admissible error region. Unlike the previous task, the mounted pose in this scenario is constrained within the SE(2) space—i.e., translations along the \(x\) and \(y\) directions and rotation about the yaw axis. 


\subsubsection{Optimization results}
\begin{figure*}[h]
  \centering
\includegraphics[width=0.98\linewidth]{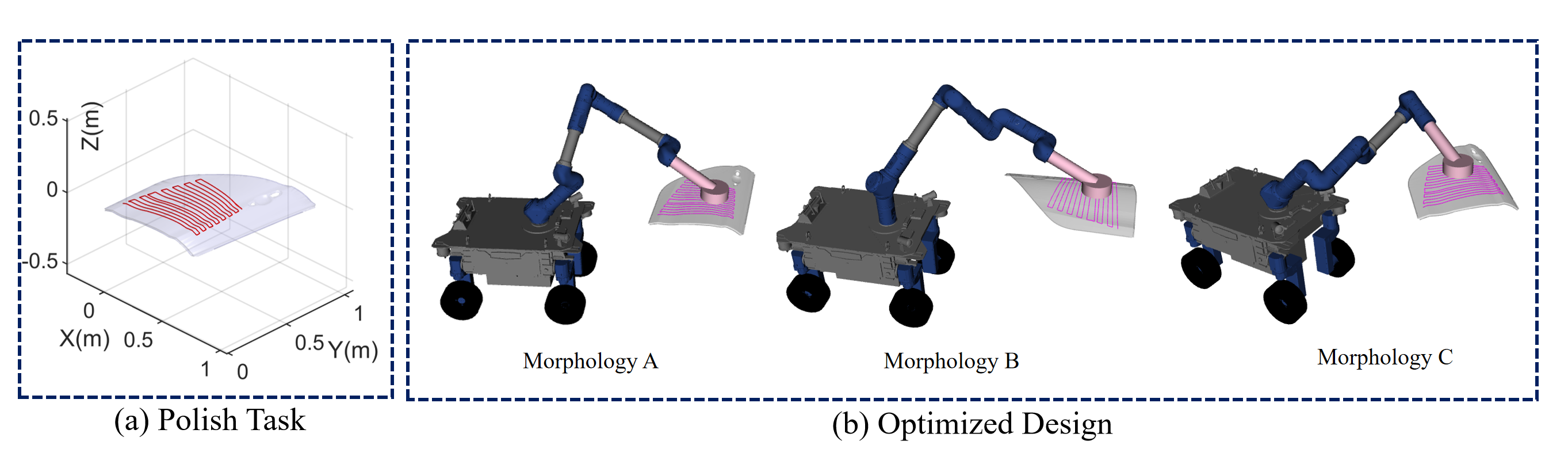}\vspace{-4mm}
  \caption{\textbf{Optimized designs and the movements for polishing task}: (Left) Polish trajectory; (Right) Morphology A denotes the morphology optimized for the combined objective of maximizing manipulability and minimizing joint effort, while Morphology B and Morphology C represent morphology optimized for maximal manipulability and minimum joint effort, respectively.  
}
  \label{fig:polish}
\end{figure*}

With the proposed design framework, we obtained three optimal morphologies for the above three scenarios, as reported in Table~\ref{optimalresult} and Fig.~\ref{fig:polish}. In particular, Morphology~A, B, and Morphology~C corresponds to the \textbf{Scenario 1}, \textbf{Scenario 2}, and \textbf{Scenario 3}, respectively.

As can be seen from Table~\ref{optimalresult}, when optimizing solely for manipulability (Morphology B with scenario 2), the 'manipulability' increases from 0.435 (Morphology~A) to 0.602, increasing by
\( 38.39\%\), and increases from 0.280 (Morphology~C) to 0.602, increasing by
\(115.00\%\). Further observation reveals that {the Morphology~B has one more DoF compared with the Morphology~A and Morphology~C, which increases the manipulator’s redundancy. The added redundancy provides greater flexibility to adjust joint configurations in the Jacobian’s null space, thereby enhancing the manipulator's manipulability.
}
{
Conversely, Morphology~C (see the third row of Table~\ref{optimalresult}), which is optimized solely for minimizing joint effort, reduces the total joint effort from 198.7 (Morphology~A) to 181.9. Compared to Morphology~B, the joint effort is reduced from 286.4 to 191.7. Unlike Morphologies~A and B, Morphology~C uses fewer modules to construct the final manipulator, lowering the overall structural load and thereby reducing the joint effort required for executing the desired task.
}

\begin{table*}[h!]
  \centering
  \renewcommand{\arraystretch}{1.25}
  \begin{tabular}{|c|c|c|c|c|c|c|}
    \hline
    \textbf{Morphology}  & \textbf{Mounted Pose} &  \textbf{Scenario}  &\textbf{Effort} & \textbf{Manipulability} & \textbf{Combined Cost} & \textbf{Total Modules} \\
    \hline
    {A (Simulation)} 
        & \multirow{2}{*}{[-1.20, 0.33, 0.15]} & \multirow{2}{*}{1} & 198.7 & 0.435  & \textbf{0.163} & \multirow{2}{*}{9 (6 DoFs) }\\
     {A (Experiment)}    & &  & 201.7 & 0.425 & \textbf{0.108} & \\
    \cline{1-7}
    {B (Simulation)} 
        & \multirow{2}{*}{[0.13, 0.09, 1.57]} & \multirow{2}{*}{2} & 286.4 & \textbf{0.602}  & 0.146 & \multirow{2}{*}{9 (7 DoFs)} \\
    {B (Experiment)}     & & & 296.7  & \textbf{0.612}  & 0.093 & \\
    \cline{1-7}
    {C (Simulation)}
        & \multirow{2}{*}{[0.29, -0.03, 2.23]} & \multirow{2}{*}{3} & \textbf{181.9}  & 0.280  & -0.419 & \multirow{2}{*}{8 (6 DoFs)} \\
    {C (Experiment)}    & &  & \textbf{191.7} & 0.293 & -0.452 & \\
    \hline
  \end{tabular}
\caption{\textbf{Optimized results for the polishing task.} The combined cost—used as the optimization objective for selecting the final experimental morphology—was computed by \(-w_f F_{\text{eff}} + w_m M_{\text{man}}\) where \(w_f = 0.01\) and \(w_m = 5\).}
  \label{optimalresult}
\end{table*}
\subsubsection{Comparison study}
\label{subsec:warmstart}
\paragraph{\textit{1) Baseline setup}}

We compare our CMA-ES approach with a baseline method proposed in the existing work~\citep{romiti2023optimization}.
{In the prior work, the mounted pose was discretized into a predefined set of candidate states, transforming its selection into a discrete search problem. However, that work only accounts for the single-branch morphology optimization. \textcolor{black}{To make a fair comparison, we here also select the mounting hole order with \( \bm S \).} As a result, both the mounted pose and morphology were encoded in a discrete optimized space and jointly solved as a combinatorial optimization problem using genetic algorithms (GAs). 

{
{In this baseline, the feasible mounted pose is discretized within a workspace where both \(x\) and \(y\) coordinates range from \(-1.2\,\text{m}\) to \(1.2\,\text{m}\) with a step size of \(0.2\,\text{m}\). The base orientation around the world-frame \(z\)-axis is discretized over \([-\pi, \pi]\) with a resolution of \(\frac{\pi}{2}\). This discretization converts the continuous mounted pose space into a finite set of candidate configurations, enabling the selection of both the modular arrangement and the mounted pose from the predefined candidate discrete states. 
The population size is set to 40, with a mutation probability of 0.5 and a crossover probability of 0.1. The algorithm is initially set to run for 100 generations. If, within these 100 generations, the cost error between consecutive generations over the last 10 generations falls below 0.005, the process terminates early. Otherwise, the evolution continues until reaches 200 generations.
}
}




\paragraph{\textit{2) Results}}
{Fig.~\ref{fig:cmaes_compare} illustrates the convergence process by displaying the minimum optimized cost value among all sampled populations at each generation. A cost value below zero indicates that the design solution satisfies all constraints defined in Eq.~\eqref{eq:cames_constrants}---namely, the robot avoids collisions with the environment, adheres to the manipulator's dynamic model constraints, and ensures successful task execution. As it can be seen from Fig.~\ref{fig:cmaes_compare}, CMA-ES achieves faster convergence and lower final cost values than GA.
}

{Assuming equal computation time per generation for CMA-ES and GA (considering they share the population size), the faster convergence of CMA-ES implies better time efficiency. In addition, the discretization used in the baseline method may exclude high-quality solutions that lie between the predefined discrete states. Although increasing the discretization resolution may improve solution quality, it also significantly enlarges the search space, leading to increased computational demands. In contrast, the CMA-ES-based optimization approach operates directly in the continuous space, circumventing the limitations introduced by discretization in the GA-based method. That is, our method enables more comprehensive and efficient exploration of the entire solution space, covering regions that would otherwise be constrained or entirely omitted by a discretized representation.
}



\begin{table*}[h]
\centering
\begin{tabular}{|c|c|c|c|c|c|c|}
\hline
\!\!\!\!\! \textbf{Morphology} \!\!\!\!\! & \textbf{Mounted Pose}  & \textbf{Effort} & \textbf{Manipulability} & \!\!\! \textbf{Combine Cost} \!\!\! & \!\!\!\!\! \textbf{Sub-branch: DoF} \!\!\!\!\! & \!\!\!\!\! \textbf{Max Torque} \!\!\!\!\! \\ \hline
A (Simulation)           & [0.05, 0.02, 0.03]        & 412.41          & 0.942                   & 0.5859     & \text{include}   : \text{0}        & \text{148 Nm}        \\ \hline
B (Simulation)           &  [0.42, 0.08, -0.44]       & 330.49          & 0.864                   & 1.0151      & \text{include}   : \text{1}      & \text{158 Nm}      \\ \hline
C (Simulation)           &   [-0.12, 0.43, 2.80]      & 398.27          & 0.988                   & 0.9573       & \text{include}   : \text{1}    & \text{148 Nm}        \\ \hline


    \cline{1-7}
    {D (Simulation)}
        & \multirow{2}{*}{ [-0.03, 0.98, 1.56] } & 431.15 & 1.082 & \textbf{1.0985}  & \multirow{2}{*}{ \text{include} :  \text{2}} & \text{152 Nm} \\
    {D (Experiment)}    & & 456.79 & 1.197 & \textbf{1.1971} &  & \text{153 Nm}  \\ \hline
\end{tabular}
\caption{\textbf{Optimized results under different initial guesses for the drilling task.} "Max Torque" represents the maximal joint torque across all joints, at all time steps. }
\label{tab:bi_branch}
\end{table*}

\begin{figure*}[ht]
    \centering
  \centering  \includegraphics[width=0.90\textwidth, trim=0pt 0pt 0pt 0pt, clip]{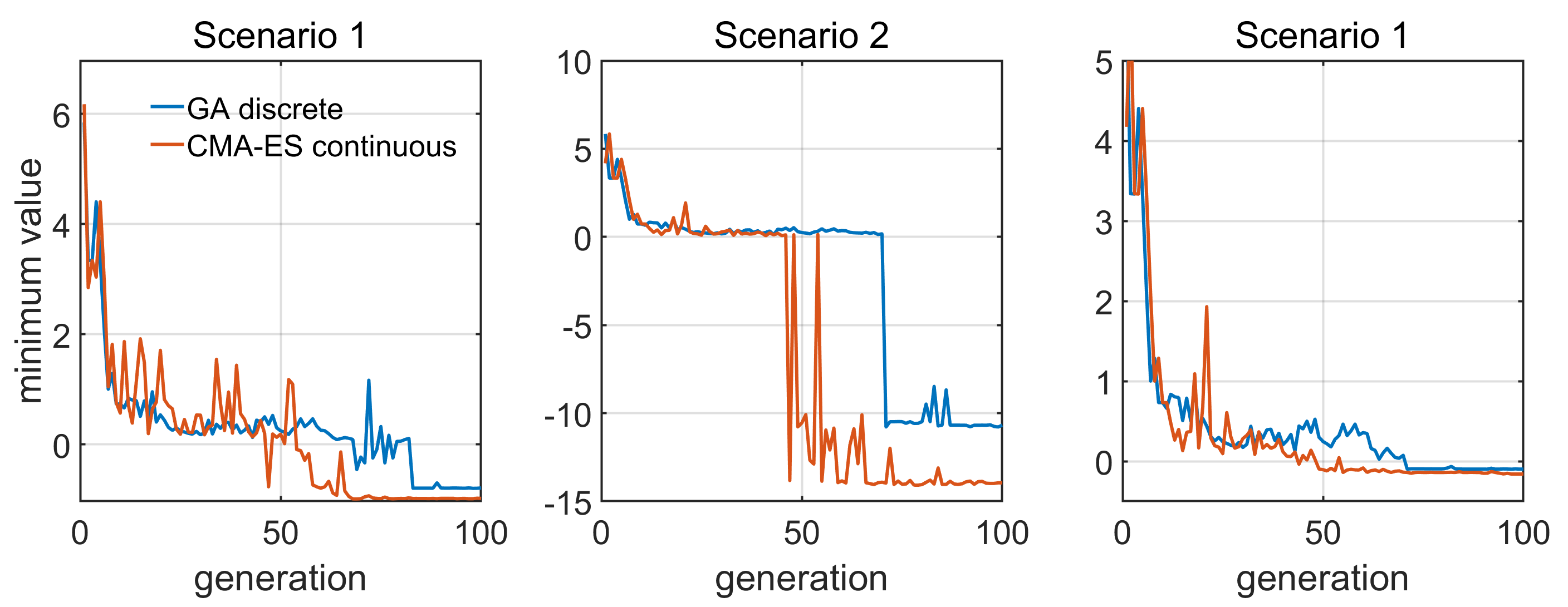}\vspace{-2mm}
\caption{\textbf{Comparison between the GA baseline and our proposed approach.} Here, `GA discrete' discretized the mounted pose into predefined states, while `CMA-ES continuous' uses a mapping function to represent discrete states in a continuous space. The curves show the evolution of the minimum cost across generations under three different optimization objectives.}
  \label{fig:cmaes_compare}
    \label{fig:cost}
\end{figure*}

    


\begin{figure*}[h]
  \centering
\includegraphics[width=0.98\textwidth]{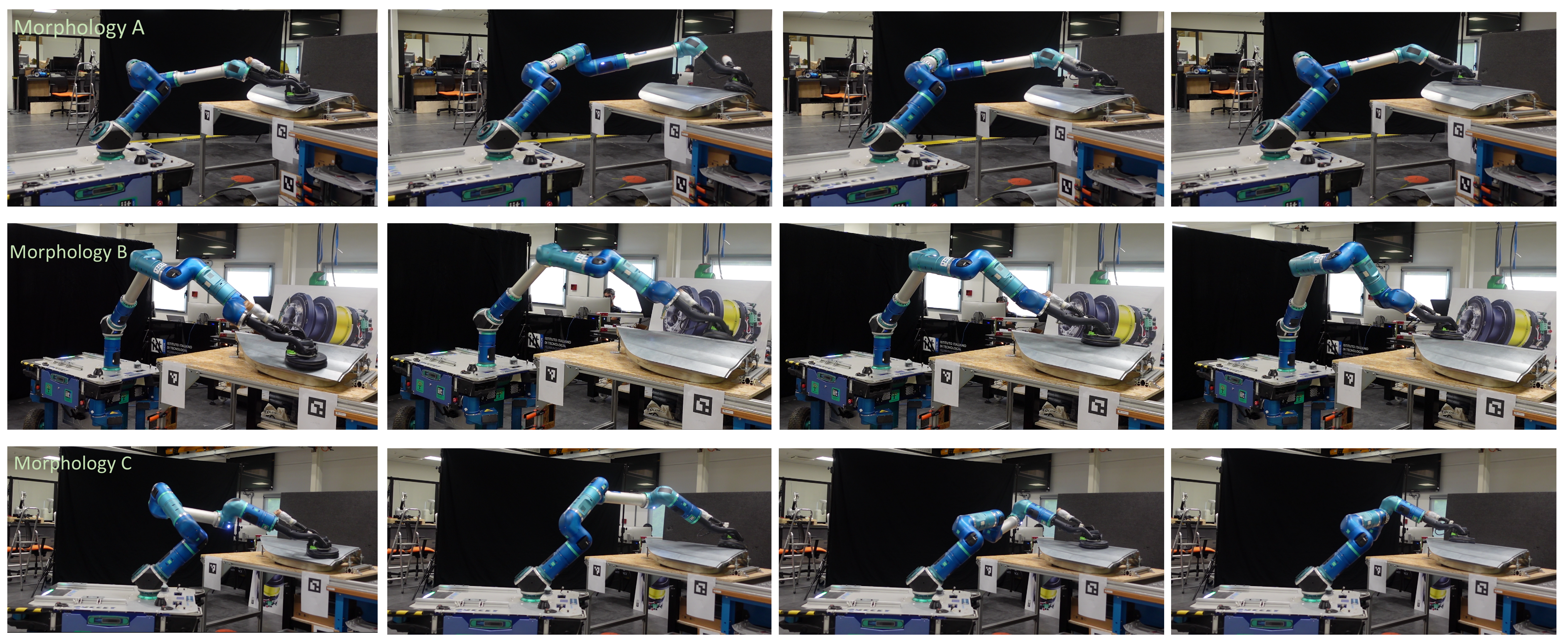}\vspace{-2mm}
  \caption{\textbf{Snapshots of hardware movements.} From top to bottom, we demonstrate results with Morphology A,  B and C.}
  \label{fig:experiments}
\end{figure*}

\begin{figure*}
  \centering
\includegraphics[width=1.0\textwidth, trim=80pt 10pt 20pt 50pt, clip]{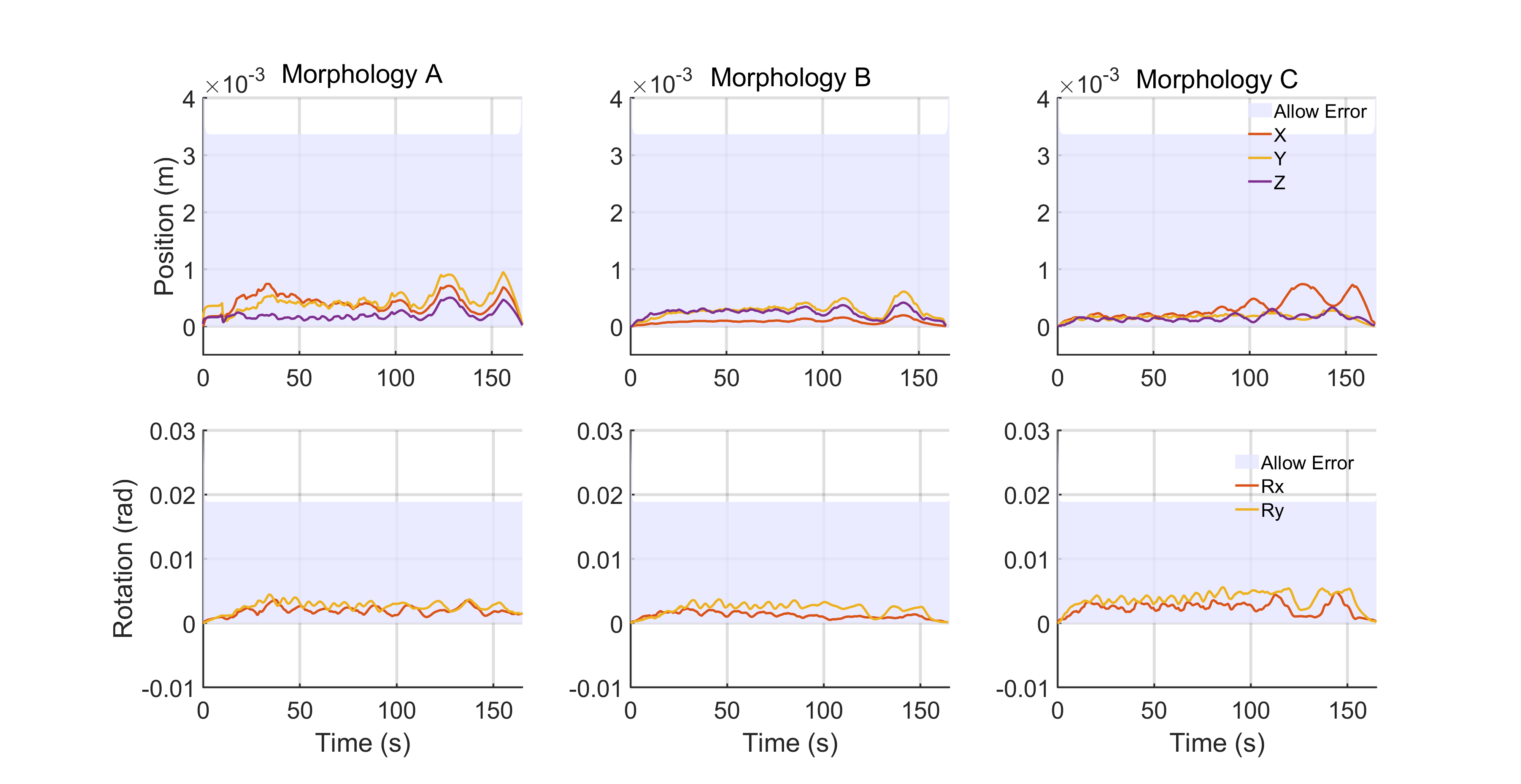}\vspace{-4mm}
  \caption{\textbf{Absolute tracking errors of the three optimal designs}. The blue shaded region represents the allowable tracking error boundary. The first row illustrates the positional tracking errors along the \(x\), \(y\), and \(z\) axes, while the second row shows the orientation tracking errors.}
  \label{fig:polish_tracking}
\end{figure*}

\subsubsection{Tracking performance in hardware experiments} 

To further validate the results, we conducted hardware experiments. 
The snapshots are shown in Fig.~\ref{fig:experiments}, where the top, middle, and bottom rows correspond to Morphology~A, Morphology~B, and Morphology~C, respectively. Unlike the pick-and-place task with only two discrete waypoints, this scenario involves 15600 waypoints. 
At each waypoint, the translational tracking error is required to be below \(0.0005\,\mathrm{m}\) and orientation errors below \(0.02\,\mathrm{rad}\).
From Fig.~\ref{fig:experiments}, we can see that three morphologies could also achieve the desired task. In addition, as shown in Fig.~\ref{fig:polish_tracking}, the tracking error in the hardware remains consistently below the specified thresholds throughout the task. Specifically, the absolute tracking-error curves lie within the allowable region, demonstrating the effectiveness of the proposed framework in real-world scenarios. {The detailed results for manipulability and joint effort metrics from the experiments are reported in Table~\ref{optimalresult}.} 
For more details, please check the attached video.

\begin{figure*}
  \centering
  \includegraphics[width=1.0\textwidth]{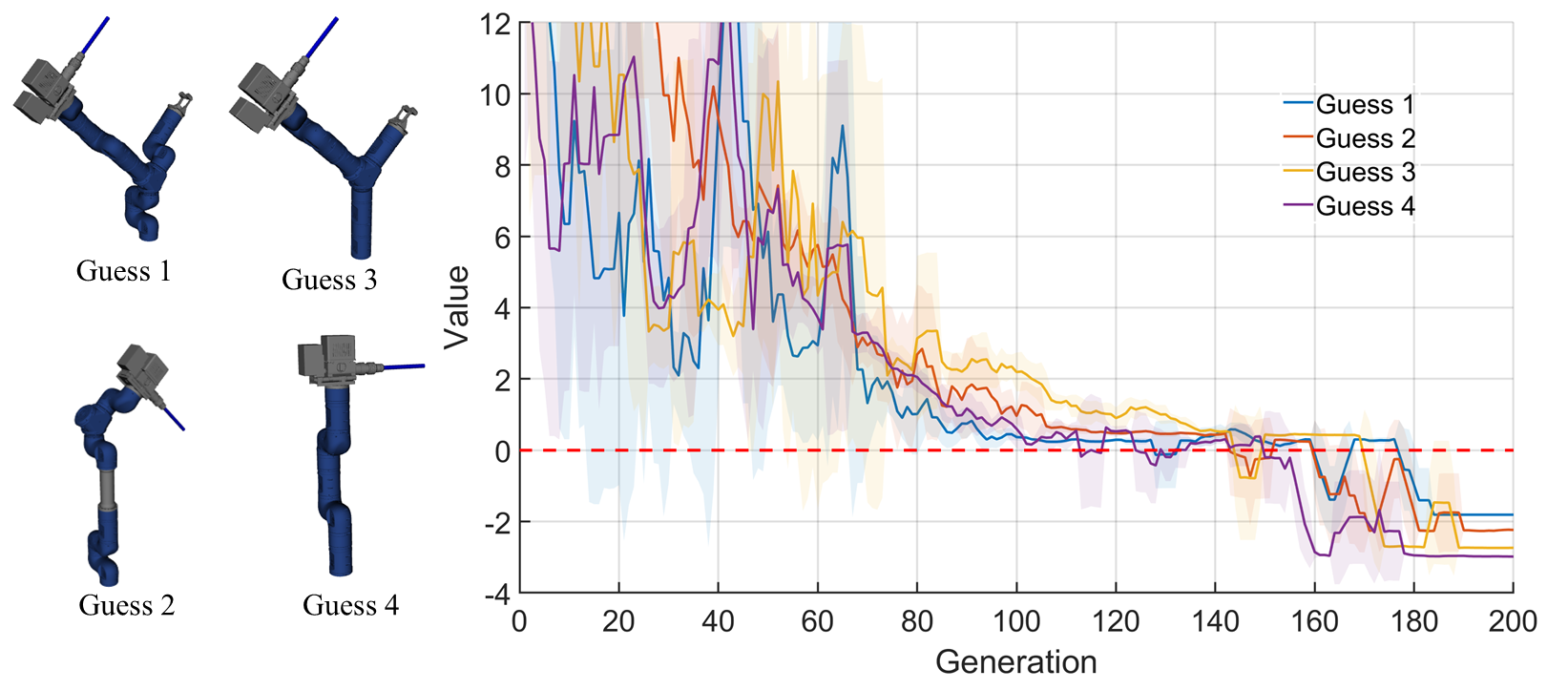}\vspace{-2mm}
  \caption{ \textbf{Optimal design process for the drilling task. } (Left) Four different initial guesses; (Right) each curve initiated with a different initial guess, where the solid lines depict the moving averages of the minimal evaluation values and the shaded areas illustrate standard deviations.}
  \label{fig:bi_branch_convergence_initially_guess}
\end{figure*}

\begin{figure*}
  \centering
  \includegraphics[width=1.0\textwidth]{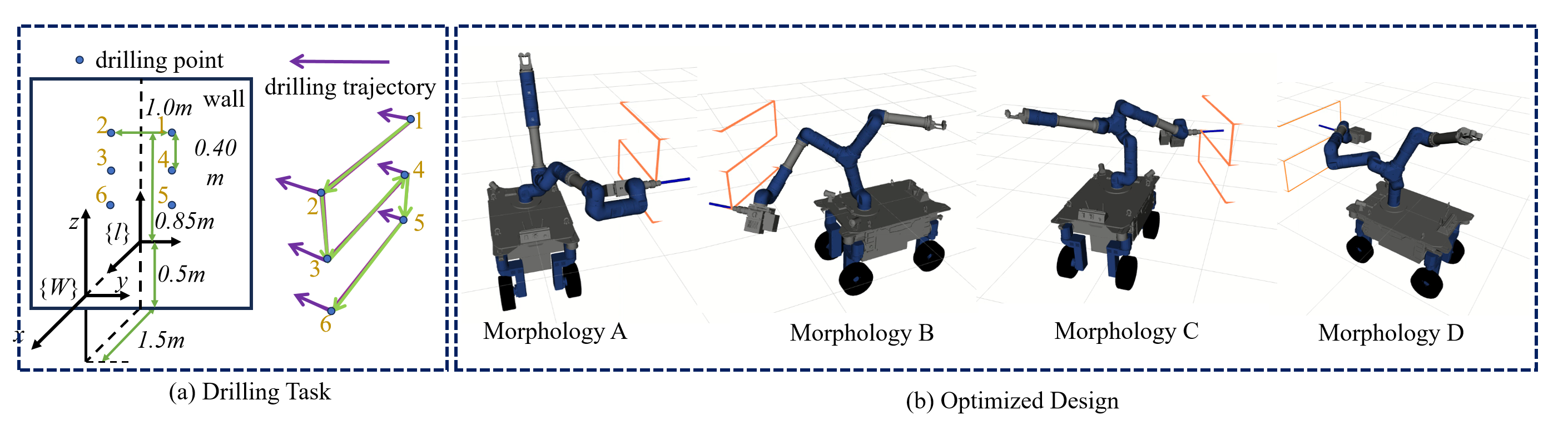}\vspace{-2mm}
  \caption{ \textbf{Drilling task and the four optimized morphologies.} (Left) Spatial arrangement of the six designated drilling targets mounted on the vertical wall, the purple line represents the drilling trajectory, the green line denotes the transportation trajectory between drilling sites; (Right) Optimized morphologies labelled A through D, corresponding respectively to guess~1 to 4.}
  \label{fig:optimal_results_bi_branch}
\end{figure*}

\begin{figure}[h]
  \centering
  \includegraphics[width=0.50\textwidth]{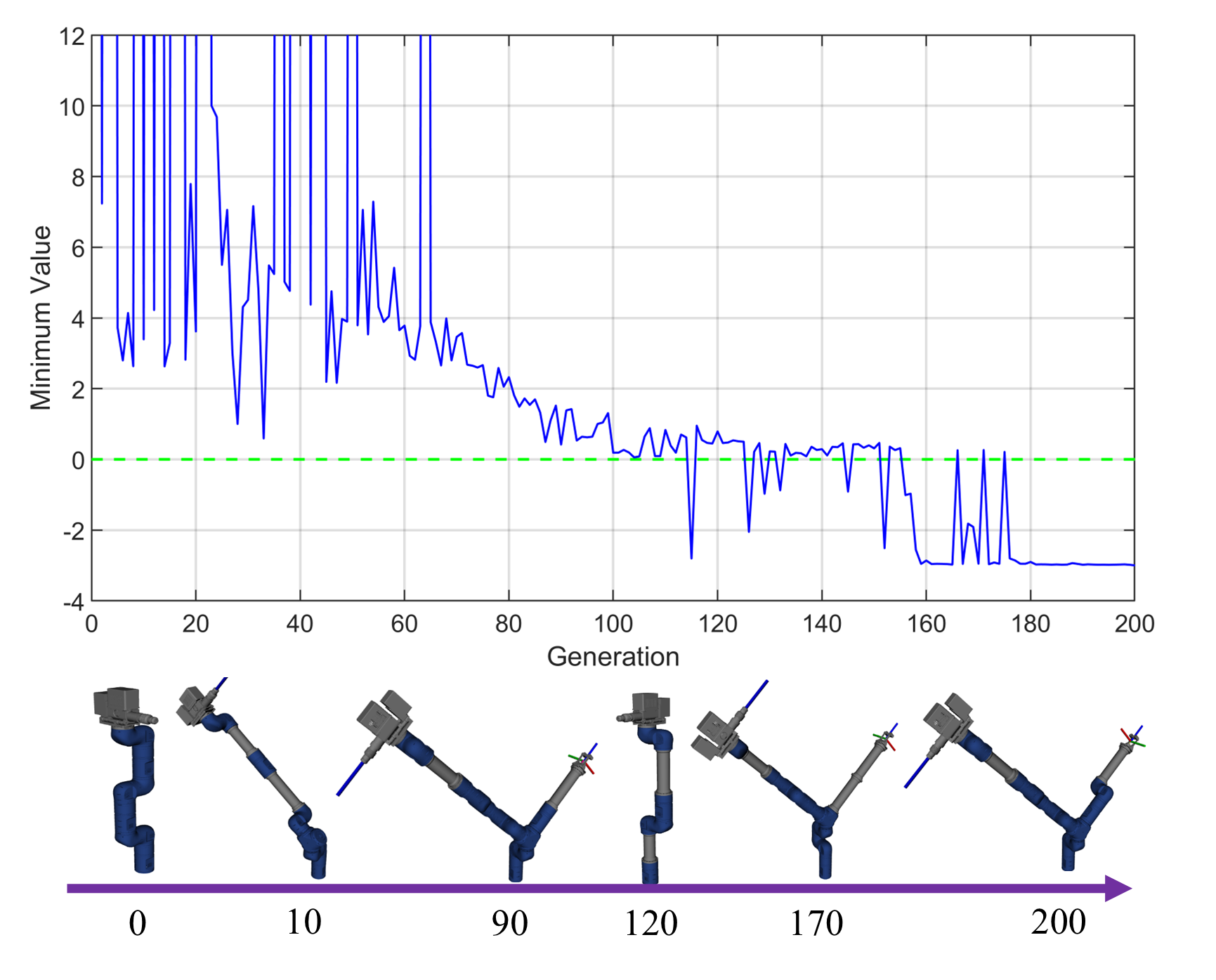}\vspace{-2mm}
  \caption{\textbf{The evolution process of Morphology~D}. The blue curve in the top panel indicates the minimal objective value among all sampled candidates at each generation, and the bottom panel visualizes the morphology evolution.}
  \label{fig:anylisis_bibranch}
\end{figure}

\begin{figure}[h]
  \centering
  \includegraphics[width=0.50\textwidth, trim=105pt 20pt 105pt 40pt, clip]{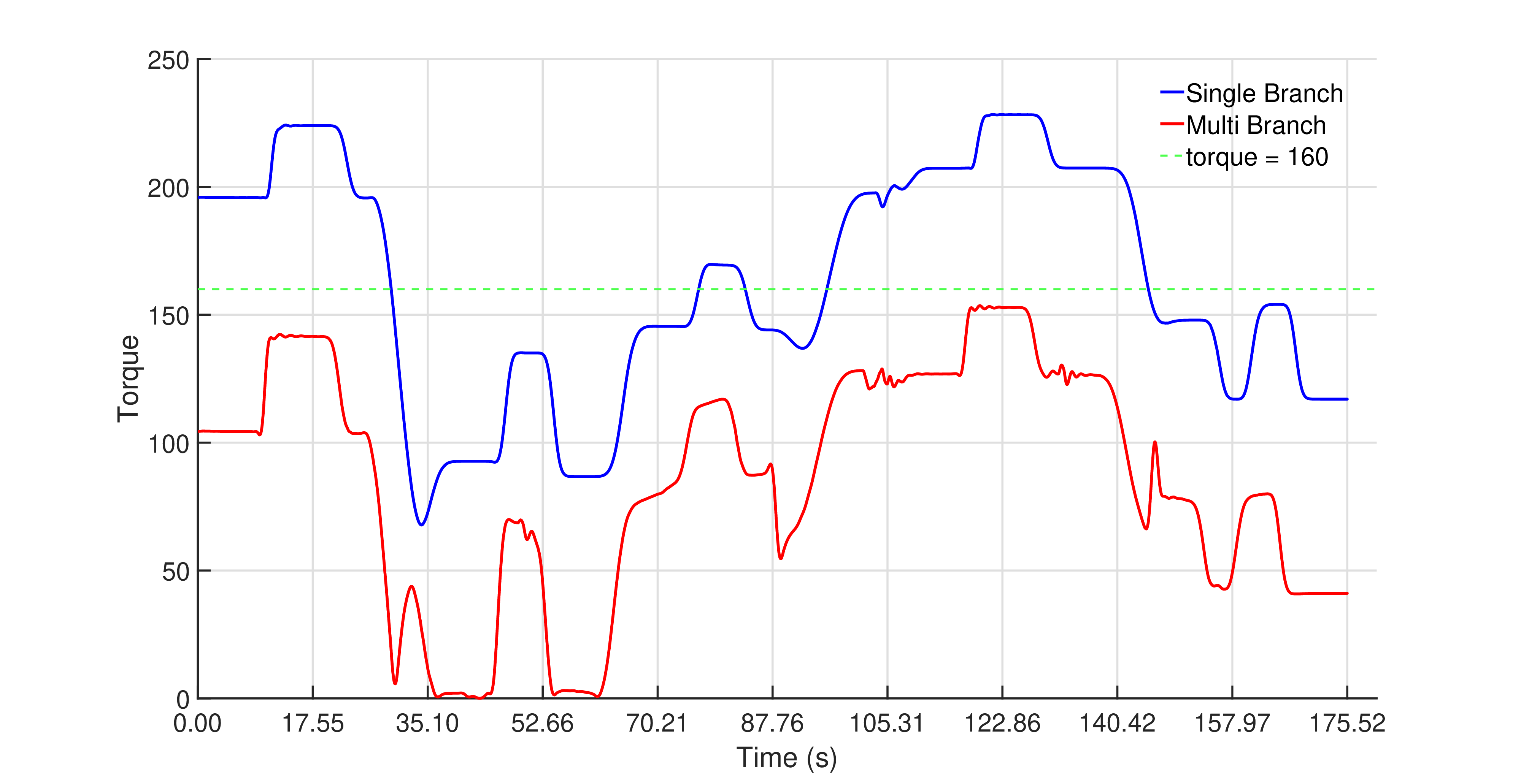}\vspace{-2mm}
  \caption{Joint torques at the common elbow joint module for bi-branch and single-branch morphologies.
}
  \label{fig:comparision_torque}
\end{figure}

\begin{figure*}
  \centering
  \includegraphics[width=0.95\textwidth]{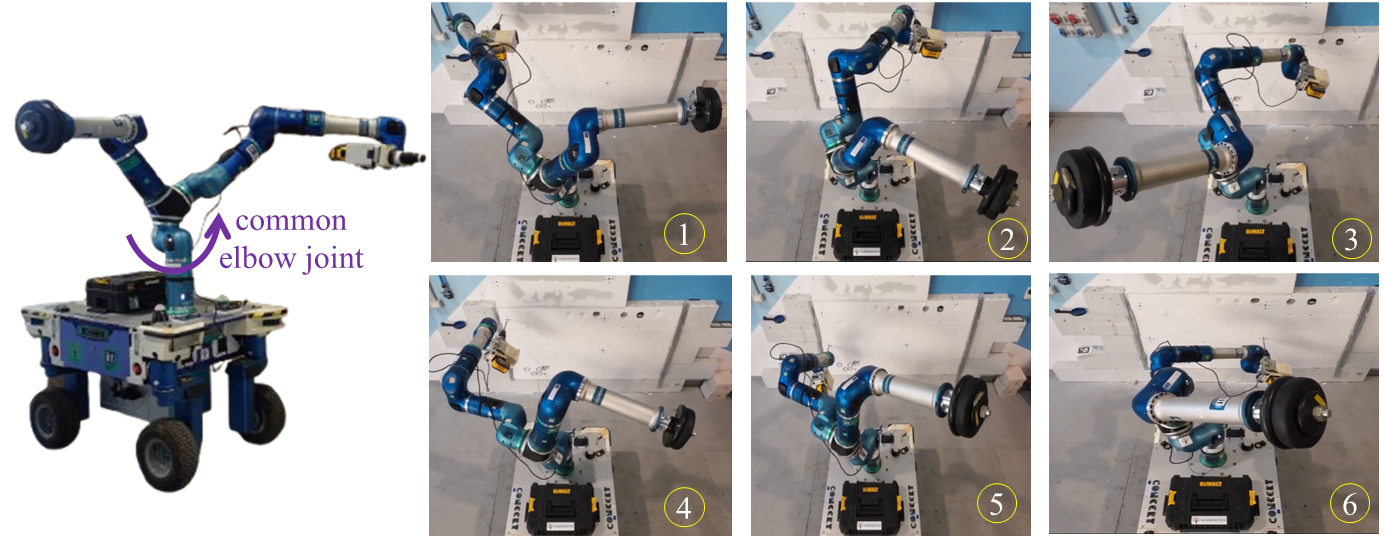}
  \caption{\textbf{Snapshot of the experimental process.} Left side illustrates the ``common elbow joint" on the bi-branch robot. The right illustrates the drilling motions from targets 1 to 6.}
  \label{fig:bi_branch_experiment}
\end{figure*}

\subsection{Drilling Task in Larger Workspace}
\subsubsection{Task description}
To further demonstrate the capability of the proposed optimization framework, especially when working in an extended workspace, we conducted a comparative study with an expanded drilling task. Compared to the previous work~\citep{lei2024task}, we span the working area from \(0.5\,\text{m} \times 0.3\,\text{m}\) to \(1.0\,\text{m} \times 0.8\,\text{m}\), doubling the range of manipulation. 
In addition, this drilling task requires the manipulator to reach six predefined locations on a vertical wall, by maintaining an orientation perpendicular to the wall surface at each target. In this scenario, rotation around the drill bit axis, i.e., end-effector \(z\) axis, is considered as a secondary-priority subtask in the HMPC formulation.

In this task, the optimization objectives in Eq.~\eqref{eq:cmaesopt} are set to minimize joint effort and maximize manipulability, with \(w_f = 0.01\) and \(w_m = 5\), respectively. In addition, the mounted pose is restricted within the SE(2) space, i.e., translations along the \(x\) and \(y\) axes and rotation about the yaw axis, since the manipulator is mounted on a mobile platform that supports only planar repositioning. To avoid collisions, the following constraints are additionally considered: (i) avoiding contact with the mobile robot platform, and (ii) avoiding collision with environmental obstacles. 
The layout of the six drilling targets is illustrated in Fig.~\ref{fig:optimal_results_bi_branch} (left), where dashed lines indicate spatial distances and solid arrows denote local coordinate frames. The right panel within the Fig.~\ref{fig:optimal_results_bi_branch} (left) presents the corresponding 3D trajectory: purple segments represent linear drilling motions with a penetration depth of \(0.15\,\text{m}\), while green segments represent transition motions between neighbouring drilling targets.

\subsubsection{Optimization with different initial guesses}
In this task, we started with four initial guesses, including single-branch and bi-branch morphologies, as shown on the left side of Fig.~\ref{fig:bi_branch_convergence_initially_guess}. 
The best fitness value at each generation is plotted on the right side of Fig.~\ref{fig:bi_branch_convergence_initially_guess}. 
Differing from the pick-and-place task and polishing task that converge after 100 generations, the drilling task required 200 generations.
From Fig.~\ref{fig:bi_branch_convergence_initially_guess}, we can find that, after 180 generations, the fitness values with the four initial guesses all drop below zeros, indicating that the physical constraints are satisfied.

The resultant optimal morphologies are illustrated on the right side of Fig.~\ref{fig:optimal_results_bi_branch} and the metrics are summarized in Table~\ref{tab:bi_branch}. We can find that the CMA-ES converged to bi-branch solutions despite being initialized with single morphologies, e.g., Guess 2 and Guess 4. This is because we strictly enforce the torque limits in the problem formulation. 

As detailed in Table~\ref{tab:bi_branch}, the assist branches exhibit 0, 1, 1, and 2 DoFs, respectively (Simulated behaviour with four morphologies can be seen in the attached video). In particular, Morphology A features a passive assist branch with 0 DoF that maintains a fixed CoM throughout the task. Despite having no active joint adjustment, the assist branch still effectively reduces the torque required at the base joints, highlighting the advantage of without needing for manufacture the new modules. 
Furthermore, Morphology~D, featuring an assist branch with two DoFs, performs best with the lowest `combine cost'. 

To further investigate the result with Morphology D, we visualize the iteration process between generations in Fig.~\ref{fig:anylisis_bibranch}. As illustrated at the bottom of Fig.~\ref{fig:anylisis_bibranch}, the initial morphology at generation~0 was a single-branch manipulator. By generation~10, the morphology has extended in length by adding modules to reach the desired pose. By generation~90, a second branch has emerged, redistributing the load and reducing the torque demands on the base. After that, although single-branch morphology occasionally appeared—such as at generation~120—the optimization process favoured the bi-branch structure. After generation~120, the bi-branch morphology was progressively refined, culminating in the final design at generation~200.


\subsubsection{Hardware experiment with a comparison study}

We conducted hardware tests with Morphology~D, which is outperforming among all the optimized designs, to further illustrate the benefits of the bi-branch design in reducing torque loads at the proximal base joints. Specifically, we evaluated its performance with two morphologies: (i) the optimized bi-branch manipulator in the real experiment, and (ii) the single-branch version by removing the assist branch. The hardware test with the bi-branch Morphology D is illustrated in Fig.~\ref{fig:bi_branch_experiment} and the absolute joint torque \(|\tau|\) at the ``common elbow joint" (see Fig.~\ref{fig:bi_branch_experiment}, left) during the experimental process is plotted in Fig.~\ref{fig:comparision_torque}. 

As can be seen in Fig.~\ref{fig:comparision_torque}, with the bi-branch morphology, the maximal torque remains below the limit (marked by the green dashed line). In contrast, the torque of the ``common elbow joint" of the single-branch morphology exceeds the threshold at several points. 
That is, without the assist branch, the elbow joint becomes overburdened, making it difficult to satisfy dynamic constraints. This analysis demonstrates that the bi-branch morphology is not merely an alternative solution, but a functionally advantageous design that satisfies the physical requirements in large-scale workspaces.


The experimental snapshots during task execution are presented in Fig.~\ref{fig:bi_branch_experiment} (right), which captures the drilling process from the first to the final target point. During execution, the posture of the assist branch is actively adjusted in real time. This adjustment dynamically repositions the CoM of the assist branch, reducing torque at the shared elbow joint and improving load distribution. It should be mentioned that the pose of the mobile platform is in agreement with the planned mounted pose of the base link of the manipulator. At the beginning of execution, the end-effector is aligned with the first target point on the planned trajectory. Throughout the experiment, all joint torques remained within the torque limits, and the robot successfully avoided collisions with both the mobile platform and the surrounding environment.


\section{Conclusion}~\label{conclusion}

This work presents a generalized and practical computational design framework for modular manipulators that unifies motion planning and design optimization to generate task-driven motions and optimal morphologies concurrently. The framework integrates an HMPC-based trajectory planner with a morphology optimizer. The motion planner is designed to accommodate both redundant and non-redundant manipulators. By adopting a redundancy-aware formulation, it provides flexibility in resolving conflicts among multiple end-effector sub-tasks, even for manipulators with limited DoFs. Leveraging the task-specific trajectory generated by the planner, the manipulator morphology is further optimized and evaluated. To address the challenge of jointly optimizing discrete morphology configurations and continuous mounted poses, we introduce a sorting-based mapping function that embeds discrete morphologies into a continuous representation. This transformation enables the use of gradient-free continuous optimization techniques, significantly improving convergence speed and increasing the likelihood of identifying high-quality solutions compared to traditional discrete combinatorial methods. 

To ensure dynamics feasibility of the resultant design, we incorporate practical physical constraints into the optimization process, including bounding tracking accuracy, avoiding collision, and adherence to dynamic model limitations. To achieve decent performance, we adopt a redundancy-aware HMPC formulation, ensuring that the resultant design maintains sufficient flexibility to coordinate multiple sub-tasks, even under limited DoFs. At the same time, the optimization process accounts for customized performance objectives, such as maximizing manipulability, minimizing joint effort, and reducing the number of modules, striking a balance between task performance and structural efficiency. The proposed computational design framework is validated through extensive simulations and real-world experiments on three representative tasks: pick-and-place, polishing, and drilling, each involving different end-effector tools. The validated results demonstrate that the proposed computational design framework consistently yields task-feasible, collision-free, and torque-efficient designs, while outperforming existing approaches in both solution quality and computational efficiency.

In the future,  a promising direction is to advance toward a fully integrated co-design framework, in which the controller not only guides optimal morphological design but also governs real-time control after deployment. Differing from the current work where the MPC works as a motion planner, we aim to develop a unified MPC-based controller capable of handling both single-branch and bi-branch morphologies. 
This unified real-control method would improve the adaptability and robustness of modular manipulators in complex scenarios. 

\section*{Acknowledge}~\label{acknowledge}

The authors gratefully acknowledge the support from the European Union’s Horizon Europe Research and Innovation Programme under Grant Agreement No. 101120731 (MAGICIAN).

\bibliographystyle{SageH} 
\bibliography{reference.bib} 

@inproceedings{sathuluri2023robust,
  title={Robust co-design of robots via cascaded optimisation},
  author={Sathuluri, Akhil and Sureshbabu, Anand Vazhapilli and Zimmermann, Markus},
  booktitle={2023 IEEE International Conference on Robotics and Automation (ICRA)},
  pages={11280--11286},
  year={2023},
  organization={IEEE}
}

@inproceedings{oleynikova2016signed,
  title={Signed distance fields: A natural representation for both mapping and planning},
  author={Oleynikova, Helen and Millane, Alexander and Taylor, Zachary and Galceran, Enric and Nieto, Juan and Siegwart, Roland},
  booktitle={RSS 2016 workshop: geometry and beyond-representations, physics, and scene understanding for robotics},
  year={2016},
  organization={University of Michigan}
}

@article{cursi2022optimization,
  title={Optimization of surgical robotic instrument mounting in a macro--micro manipulator setup for improving task execution},
  author={Cursi, Francesco and Bai, Weibang and Yeatman, Eric M and Kormushev, Petar},
  journal={IEEE Transactions on Robotics},
  volume={38},
  number={5},
  pages={2858--2874},
  year={2022},
  publisher={IEEE}
}

@article{hansen2003reducing,
  title={Reducing the time complexity of the derandomized evolution strategy with covariance matrix adaptation (CMA-ES)},
  author={Hansen, Nikolaus and M{\"u}ller, Sibylle D and Koumoutsakos, Petros},
  journal={Evolutionary computation},
  volume={11},
  number={1},
  pages={1--18},
  year={2003},
  publisher={MIT Press}
}

@inproceedings{qin2022install,
  title={Where to Install the Manipulator: Optimal Installation Pose Planning Based on Whale Algorithm},
  author={Qin, Xinyu and Zhang, Heng and Zhou, Tieliang and Xiong, Zhenhua},
  booktitle={2022 IEEE/ASME International Conference on Advanced Intelligent Mechatronics (AIM)},
  pages={962--967},
  year={2022},
  organization={IEEE}
}

@inproceedings{romiti2023optimization,
  title={An Optimization Study on Modular Reconfigurable Robots: Finding the Task-Optimal Design},
  author={Romiti, Edoardo and Iacobelli, Francesco and Ruzzon, Marco and Kashiri, Navvab and Malzahn, J{\"o}rn and Tsagarakis, Nikos},
  booktitle={2023 IEEE 19th International Conference on Automation Science and Engineering (CASE)},
  pages={1--8},
  year={2023},
  organization={IEEE}
}

@article{kramer2020model,
  title={Model predictive control of a collaborative manipulator considering dynamic obstacles},
  author={Kr{\"a}mer, Maximilian and R{\"o}smann, Christoph and Hoffmann, Frank and Bertram, Torsten},
  journal={Optimal Control Applications and Methods},
  volume={41},
  number={4},
  pages={1211--1232},
  year={2020},
  publisher={Wiley Online Library}
}

@article{minniti2019whole,
  title={Whole-body mpc for a dynamically stable mobile manipulator},
  author={Minniti, Maria Vittoria and Farshidian, Farbod and Grandia, Ruben and Hutter, Marco},
  journal={IEEE Robotics and Automation Letters},
  volume={4},
  number={4},
  pages={3687--3694},
  year={2019},
  publisher={IEEE}
}

@article{ferreau2014qpoases,
  title={qpOASES: A parametric active-set algorithm for quadratic programming},
  author={Ferreau, Hans Joachim and Kirches, Christian and Potschka, Andreas and Bock, Hans Georg and Diehl, Moritz},
  journal={Mathematical Programming Computation},
  volume={6},
  number={4},
  pages={327--363},
  year={2014},
  publisher={Springer}
}

@inproceedings{lei2022mpc,
  title={An MPC-Based Framework for Dynamic Trajectory Re-Planning in Uncertain Environments},
  author={Lei, Maolin and Lu, Liang and Laurenzi, Arturo and Rossini, Luca and Romiti, Edoardo and Malzahn, J{\"o}rn and Tsagarakis, Nikos G},
  booktitle={2022 IEEE-RAS 21st International Conference on Humanoid Robots (Humanoids)},
  pages={594--601},
  year={2022},
  organization={IEEE}
}

@inproceedings{lei2022dual,
  title={Dual-arm object transportation via model predictive control and external disturbance estimation},
  author={Lei, Maolin and Selvaggio, Mario and Wang, Ting and Ruggiero, Fabio and Zhou, Cheng and Yao, Chen and Zheng, Yu},
  booktitle={2022 IEEE 18th International Conference on Automation Science and Engineering (CASE)},
  pages={2328--2334},
  year={2022},
  organization={IEEE}
}

@article{huo2005kinematic,
  title={Kinematic inversion of functionally-redundant serial manipulators: application to arc-welding},
  author={Huo, Liguo and Baron, Luc},
  journal={Transactions of the Canadian Society for Mechanical Engineering},
  volume={29},
  number={4},
  pages={679--690},
  year={2005}
}

@article{mansard2009directional,
  title={Directional redundancy for robot control},
  author={Mansard, Nicolas and Chaumette, Fran{\c{c}}ois},
  journal={IEEE Transactions on Automatic Control},
  volume={54},
  number={6},
  pages={1179--1192},
  year={2009},
  publisher={IEEE}
}

@inproceedings{slotine1991general,
  title={A general framework for managing multiple tasks in highly redundant robotic systems},
  author={Slotine, Siciliano B and Siciliano, B},
  booktitle={proceeding of 5th International Conference on Advanced Robotics},
  volume={2},
  pages={1211--1216},
  year={1991}
}

@article{zanchettin2011use,
  title={On the use of functional redundancy in industrial robotic manipulators for optimal spray painting},
  author={Zanchettin, Andrea Maria and Rocco, Paolo},
  journal={IFAC Proceedings Volumes},
  volume={44},
  number={1},
  pages={11495--11500},
  year={2011},
  publisher={Elsevier}
}

@article{gupta1986nature,
  title={On the nature of robot workspace},
  author={Gupta, KC},
  journal={The International journal of robotics research},
  volume={5},
  number={2},
  pages={112--121},
  year={1986},
  publisher={Sage Publications Sage CA: Thousand Oaks, CA}
}

@inproceedings{liu2020optimizing,
  title={Optimizing performance in automation through modular robots},
  author={Liu, Stefan B and Althoff, Matthias},
  booktitle={2020 IEEE International Conference on Robotics and Automation (ICRA)},
  pages={4044--4050},
  year={2020},
  organization={IEEE}
}

@article{ha2018computational,
  title={Computational design of robotic devices from high-level motion specifications},
  author={Ha, Sehoon and Coros, Stelian and Alspach, Alexander and Bern, James M and Kim, Joohyung and Yamane, Katsu},
  journal={IEEE Transactions on Robotics},
  volume={34},
  number={5},
  pages={1240--1251},
  year={2018},
  publisher={IEEE}
}

@article{zhao2020robogrammar,
  title={Robogrammar: graph grammar for terrain-optimized robot design},
  author={Zhao, Allan and Xu, Jie and Konakovi{\'c}-Lukovi{\'c}, Mina and Hughes, Josephine and Spielberg, Andrew and Rus, Daniela and Matusik, Wojciech},
  journal={ACM Transactions on Graphics (TOG)},
  volume={39},
  number={6},
  pages={1--16},
  year={2020},
  publisher={ACM New York, NY, USA}
}

@inproceedings{romiti2021minimum,
  title={Minimum-Effort Task-based Design Optimization of Modular Reconfigurable Robots},
  author={Romiti, Edoardo and Kashiri, Navvab and Malzahn, J{\"o}rn and Tsagarakis, Nikos},
  booktitle={2021 IEEE International Conference on Robotics and Automation (ICRA)},
  pages={9891--9897},
  year={2021},
  organization={IEEE}
}

@article{romiti2021toward,
  title={Toward a plug-and-work reconfigurable cobot},
  author={Romiti, Edoardo and Malzahn, J{\"o}rn and Kashiri, Navvab and Iacobelli, Francesco and Ruzzon, Marco and Laurenzi, Arturo and Hoffman, Enrico Mingo and Muratore, Luca and Margan, Alessio and Baccelliere, Lorenzo and others},
  journal={IEEE/ASME transactions on mechatronics},
  volume={27},
  number={5},
  pages={3219--3231},
  year={2021},
  publisher={IEEE}
}

@inproceedings{icer2017evolutionary,
  title={Evolutionary cost-optimal composition synthesis of modular robots considering a given task},
  author={Icer, Esra and Hassan, Heba A and El-Ayat, Khaled and Althoff, Matthias},
  booktitle={2017 IEEE/RSJ International Conference on Intelligent Robots and Systems (IROS)},
  pages={3562--3568},
  year={2017},
  organization={IEEE}
}

@inproceedings{hu2023glso,
  title={GLSO: Grammar-guided Latent Space Optimization for Sample-efficient Robot Design Automation},
  author={Hu, Jiaheng and Whitman, Julian and Choset, Howie},
  booktitle={Conference on Robot Learning},
  pages={1321--1331},
  year={2023},
  organization={PMLR}
}

@inproceedings{matsumaru1995design,
  title={Design and control of the modular robot system: TOMMS},
  author={Matsumaru, Takafumi},
  booktitle={Proceedings of 1995 IEEE international conference on robotics and automation},
  volume={2},
  pages={2125--2131},
  year={1995},
  organization={IEEE}
}

@article{zhang2006novel,
  title={A novel reconfigurable robot for urban search and rescue},
  author={Zhang, Houxiang and Wang, Wei and Deng, Zhicheng and Zong, Guanghua and Zhang, Jianwei},
  journal={International Journal of Advanced Robotic Systems},
  volume={3},
  number={4},
  pages={48},
  year={2006},
  publisher={SAGE Publications Sage UK: London, England}
}

@article{yim2007modular,
  title={Modular self-reconfigurable robot systems [grand challenges of robotics]},
  author={Yim, Mark and Shen, Wei-Min and Salemi, Behnam and Rus, Daniela and Moll, Mark and Lipson, Hod and Klavins, Eric and Chirikjian, Gregory S},
  journal={IEEE Robotics \& Automation Magazine},
  volume={14},
  number={1},
  pages={43--52},
  year={2007},
  publisher={IEEE}
}

@article{raina2021impact,
  title={Impact modeling and reactionless control for post-capturing and maneuvering of orbiting objects using a multi-arm space robot},
  author={Raina, Deepak and Gora, Sunil and Maheshwari, Dheeraj and Shah, Suril V},
  journal={Acta Astronautica},
  volume={182},
  pages={21--36},
  year={2021},
  publisher={Elsevier}
}

@article{whitman2018task,
  title={Task-specific manipulator design and trajectory synthesis},
  author={Whitman, Julian and Choset, Howie},
  journal={IEEE Robotics and Automation Letters},
  volume={4},
  number={2},
  pages={301--308},
  year={2018},
  publisher={IEEE}
}

@article{du2024learning,
  title={Learning-based Multimodal Control for a Supernumerary Robotic System in Human-Robot Collaborative Sorting},
  author={Du, Yuwei and Amor, Heni Ben and Jin, Jing and Wang, Qiang and Ajoudani, Arash},
  journal={IEEE Robotics and Automation Letters},
  year={2024},
  publisher={IEEE}
}

@misc{sahinidis2019mixed,
  title={Mixed-integer nonlinear programming 2018},
  author={Sahinidis, Nikolaos V},
  journal={Optimization and Engineering},
  volume={20},
  pages={301--306},
  year={2019},
  publisher={Springer}
}

@article{krause2016cma,
  title={CMA-ES with optimal covariance update and storage complexity},
  author={Krause, Oswin and Arbon{\`e}s, D{\'\i}dac Rodr{\'\i}guez and Igel, Christian},
  journal={Advances in neural information processing systems},
  volume={29},
  year={2016}
}

@article{wang2024hierarchical,
  title={Hierarchical Incremental MPC for Redundant Robots: A Robust and Singularity-Free Approach},
  author={Wang, Yongchao and Liu, Yang and Leibold, Marion and Buss, Martin and Lee, Jinoh},
  journal={IEEE Transactions on Robotics},
  year={2024},
  publisher={IEEE}
}

@inproceedings{lee2023real,
  title={Real-time model predictive control for industrial manipulators with singularity-tolerant hierarchical task control},
  author={Lee, Jaemin and Seo, Mingyo and Bylard, Andrew and Sun, Robert and Sentis, Luis},
  booktitle={2023 IEEE International Conference on Robotics and Automation (ICRA)},
  pages={12282--12288},
  year={2023},
  organization={IEEE}
}

@article{nubert2020safe,
  title={Safe and fast tracking on a robot manipulator: Robust mpc and neural network control},
  author={Nubert, Julian and K{\"o}hler, Johannes and Berenz, Vincent and Allg{\"o}wer, Frank and Trimpe, Sebastian},
  journal={IEEE Robotics and Automation Letters},
  volume={5},
  number={2},
  pages={3050--3057},
  year={2020},
  publisher={IEEE}
}

@article{nicolis2020operational,
  title={Operational space model predictive sliding mode control for redundant manipulators},
  author={Nicolis, Davide and Allevi, Fabio and Rocco, Paolo},
  journal={IEEE Transactions on Robotics},
  volume={36},
  number={4},
  pages={1348--1355},
  year={2020},
  publisher={IEEE}
}

@inproceedings{qin1997overview,
  title={An overview of industrial model predictive control technology},
  author={Qin, S Joe and Badgwell, Thomas A},
  booktitle={AIche symposium series},
  volume={93},
  number={316},
  pages={232--256},
  year={1997},
  organization={New York, NY: American Institute of Chemical Engineers, 1971-c2002.}
}

@article{mayne2014model,
  title={Model predictive control: Recent developments and future promise},
  author={Mayne, David Q},
  journal={Automatica},
  volume={50},
  number={12},
  pages={2967--2986},
  year={2014},
  publisher={Elsevier}
}

@article{kohler2018nonlinear,
  title={Nonlinear reference tracking: An economic model predictive control perspective},
  author={K{\"o}hler, Johannes and M{\"u}ller, Matthias A and Allg{\"o}wer, Frank},
  journal={IEEE Transactions on Automatic Control},
  volume={64},
  number={1},
  pages={254--269},
  year={2018},
  publisher={IEEE}
}

@article{bouyarmane2017weight,
  title={On weight-prioritized multitask control of humanoid robots},
  author={Bouyarmane, Karim and Kheddar, Abderrahmane},
  journal={IEEE Transactions on Automatic Control},
  volume={63},
  number={6},
  pages={1632--1647},
  year={2017},
  publisher={IEEE}
}

@article{gafur2021dynamic,
  title={Dynamic collision avoidance for multiple robotic manipulators based on a non-cooperative multi-agent game},
  author={Gafur, Nigora and Kanagalingam, Gajanan and Ruskowski, Martin},
  journal={arXiv preprint arXiv:2103.00583},
  year={2021}
}

@inproceedings{gaertner2021collision,
  title={Collision-free MPC for legged robots in static and dynamic scenes},
  author={Gaertner, Magnus and Bjelonic, Marko and Farshidian, Farbod and Hutter, Marco},
  booktitle={2021 IEEE International Conference on Robotics and Automation (ICRA)},
  pages={8266--8272},
  year={2021},
  organization={IEEE}
}

@article{kim1987visual,
  title={Visual enhancements in pick-and-place tasks: Human operators controlling a simulated cylindrical manipulator},
  author={Kim, Won and Tendick, Frank and Stark, LAWRENCEW},
  journal={IEEE Journal on Robotics and Automation},
  volume={3},
  number={5},
  pages={418--425},
  year={1987},
  publisher={IEEE}
}

@article{ogbemhe2015towards,
  title={Towards achieving a fully intelligent robotic arc welding: a review},
  author={Ogbemhe, John and Mpofu, Khumbulani},
  journal={Industrial Robot: An International Journal},
  volume={42},
  number={5},
  pages={475--484},
  year={2015},
  publisher={Emerald Group Publishing Limited}
}

@article{kah2015robotic,
  title={Robotic arc welding sensors and programming in industrial applications},
  author={Kah, Paul and Shrestha, Manish and Hiltunen, Esa and Martikainen, Jukka},
  journal={International Journal of Mechanical and Materials Engineering},
  volume={10},
  pages={1--16},
  year={2015},
  publisher={Springer}
}

@article{kharidege2017practical,
  title={A practical approach for automated polishing system of free-form surface path generation based on industrial arm robot},
  author={Kharidege, Ahmed and Ting, Du Ting and Yajun, Zhang},
  journal={The International Journal of Advanced Manufacturing Technology},
  volume={93},
  pages={3921--3934},
  year={2017},
  publisher={Springer}
}

@article{xu2017kinematics,
  title={Kinematics analysis of a hybrid manipulator for computer controlled ultra-precision freeform polishing},
  author={Xu, Peng and Cheung, Chi-Fai and Li, Bing and Ho, Lai-Ting and Zhang, Ju-Fan},
  journal={Robotics and Computer-Integrated Manufacturing},
  volume={44},
  pages={44--56},
  year={2017},
  publisher={Elsevier}
}

@article{goodrich2008human,
  title={Human--robot interaction: a survey},
  author={Goodrich, Michael A and Schultz, Alan C and others},
  journal={Foundations and Trends{\textregistered} in Human--Computer Interaction},
  volume={1},
  number={3},
  pages={203--275},
  year={2008},
  publisher={Now Publishers, Inc.}
}

@article{murphy2010human,
  title={Human--robot interaction},
  author={Murphy, Robin R and Nomura, Tatsuya and Billard, Aude and Burke, Jennifer L},
  journal={IEEE robotics \& automation magazine},
  volume={17},
  number={2},
  pages={85--89},
  year={2010},
  publisher={IEEE}
}

@article{lei2024task,
  title={Task-Driven Computational Framework for Simultaneously Optimizing Design and Mounted Pose of Modular Reconfigurable Manipulators},
  author={Lei, Maolin and Romiti, Edoardo and Laurenz, Arturo and Tsagarakis, Nikos G},
  journal={arXiv preprint arXiv:2405.01923},
  year={2024}
}

@article{mayer2025holistic,
  title={Holistic Optimization of Modular Robots},
  author={Mayer, Matthias and Althoff, Matthias},
  journal={arXiv preprint arXiv:2505.00400},
  year={2025}
}

@article{rossini2025concert,
  title={CONCERT: a Modular Reconfigurable Robot for Construction},
  author={Rossini, Luca and Romiti, Edoardo and Laurenzi, Arturo and Ruscelli, Francesco and Ruzzon, Marco and Covizzi, Luca and Baccelliere, Lorenzo and Carrozzo, Stefano and Terzer, Michael and Magri, Marco and others},
  journal={arXiv preprint arXiv:2504.04998},
  year={2025}
}

@article{kennel2024payload,
  title={Payload-aware trajectory optimisation for non-holonomic mobile multi-robot manipulation with tip-over avoidance},
  author={Kennel-Maushart, Florian and Coros, Stelian},
  journal={IEEE Robotics and Automation Letters},
  year={2024},
  publisher={IEEE}
}

@article{gill2005snopt,
  title={SNOPT: An SQP algorithm for large-scale constrained optimization},
  author={Gill, Philip E and Murray, Walter and Saunders, Michael A},
  journal={SIAM review},
  volume={47},
  number={1},
  pages={99--131},
  year={2005},
  publisher={SIAM}
}

@article{unser1993b,
  title={B-spline signal processing. I. Theory},
  author={Unser, Michael and Aldroubi, Akram and Eden, Murray},
  journal={IEEE transactions on signal processing},
  volume={41},
  number={2},
  pages={821--833},
  year={1993},
  publisher={IEEE}
}

@book{williams2006gaussian,
  title={Gaussian processes for machine learning},
  author={Williams, Christopher KI and Rasmussen, Carl Edward},
  volume={2},
  number={3},
  year={2006},
  publisher={MIT press Cambridge, MA}
}

@article{siciliano1990kinematic,
  title={Kinematic control of redundant robot manipulators: A tutorial},
  author={Siciliano, Bruno},
  journal={Journal of intelligent and robotic systems},
  volume={3},
  pages={201--212},
  year={1990},
  publisher={Springer}
}

@article{koike2023simultaneous,
  title={Simultaneous optimization of discrete and continuous parameters defining a robot morphology and controller},
  author={Koike, Ryosuke and Ariizumi, Ryo and Matsuno, Fumitoshi},
  journal={IEEE Transactions on Neural Networks and Learning Systems},
  year={2023},
  publisher={IEEE}
}

@inproceedings{kulz2024optimizing,
  title={Optimizing modular robot composition: A lexicographic genetic algorithm approach},
  author={K{\"u}lz, Jonathan and Althoff, Matthias},
  booktitle={2024 IEEE International Conference on Robotics and Automation (ICRA)},
  pages={16752--16758},
  year={2024},
  organization={IEEE}
}

@inproceedings{shiller1985optimal,
  title={On the optimal control of robotic manipulators with actuator and end-effector constraints},
  author={Shiller, Zvi and Dubowsky, Steven},
  booktitle={Proceedings. 1985 IEEE International Conference on Robotics and Automation},
  volume={2},
  pages={614--620},
  year={1985},
  organization={IEEE}
}

@book{chenier1996shortest,
  title={Shortest paths in weighted polygons.},
  author={Ch{\'e}nier, Christian},
  year={1996},
  publisher={University of Ottawa (Canada)}
}

@article{holkar2010overview,
  title={An overview of model predictive control},
  author={Holkar, Kailas S and Waghmare, Laxman M},
  journal={International Journal of control and automation},
  volume={3},
  number={4},
  pages={47--63},
  year={2010}
}

@misc{hansen2019pycma,
  author       = {Nikolaus Hansen and Youhei Akimoto and Petr Baudis},
  title        = {{CMA-ES/pycma} on {G}ithub},
  howpublished = {Zenodo, DOI:10.5281/zenodo.2559634},
  month        = feb,
  year         = 2019,
  doi          = {10.5281/zenodo.2559634},
  url          = {https://doi.org/10.5281/zenodo.2559634},
}

@article{wang2021optimal,
  title={Optimal order pick-and-place of objects in cluttered scene by a mobile manipulator},
  author={Wang, Fengyi and Olvera, J Rogelio Guadarrama and Cheng, Gordon},
  journal={IEEE Robotics and Automation Letters},
  volume={6},
  number={4},
  pages={6402--6409},
  year={2021},
  publisher={IEEE}
}

@book{casella2024statistical,
  title={Statistical inference},
  author={Casella, George and Berger, Roger},
  year={2024},
  publisher={Chapman and Hall/CRC}
}
\section*{Appendix A}

\subsection*{GPR-based tracking bound interpolation
}~\label{sec:gp_fitness}


With the desired boundary at the specific way-points, the Gaussian Process Regression (GPR) can be employed for interpolating the tolerance boundary for the whole trajectory. At each specific waypoint, we model the tracking error using a zero-mean Gaussian distribution, where the covariance is set to \( \frac{ \bm{\xi} (t) }{2} \). This choice reflects the assumption that the upper and lower bounds correspond to a 95\% confidence interval. Given these per-waypoint distributions, we can construct a Gaussian distribution for all the key way-point tolerance Gaussian distributions, written as \(\boldsymbol{y} \sim \mathcal{N}(0, \text{Cov}(\bm{y}) ) \) where the covariance of the noisy observation vector is defined as
\begin{equation}
\operatorname{Cov}(\boldsymbol{y}) = \mathbf{K}(\boldsymbol{t}, \boldsymbol{t}) + \boldsymbol{\Sigma},
\end{equation}
where \(\boldsymbol{\Sigma} = \operatorname{diag}(\frac{\boldsymbol{\xi}}{2})\) represents the observation noise, and \(\mathbf{K} \in \mathbb{R}^{N_t \times N_t}\) is the kernel matrix computed using the radial basis function (RBF) kernel:
\begin{equation}
k(t_i, t_j) = \sigma_f^2 \exp\left(-\frac{(t_i - t_j)^2}{l^2} \right),
\end{equation}
where \(\sigma_f^2\) is the signal variance and \(l\) is the characteristic length scale. \(N_t\) represents the total number of key waypoints. 
We set \(\sigma_f^2 = 0.005\) for position and \(\sigma_f^2 = 0.015\) for orientation, while choosing a relatively large length scale \(l = 1.0\) to promote smooth transitions between waypoints.

According to the gaussian distribution at the specific way-points, the function values at an additional set of \(N_p\) time points \(\boldsymbol{t}_* = [t_1^*, \dots, t_{N_t}^*]^\top\), are jointly expressed as

\begin{equation}
\begin{bmatrix}
\boldsymbol{y} \\
\boldsymbol{f}_*
\end{bmatrix}
\sim \mathcal{N} \left(
\begin{bmatrix}
\boldsymbol{0} \\
\boldsymbol{0}
\end{bmatrix},
\begin{bmatrix}
\mathbf{K} + \boldsymbol{\Sigma} & \mathbf{K}_*^\top \\
\mathbf{K}_* & \mathbf{K}_{**}
\end{bmatrix}
\right),
\end{equation}
where \(\mathbf{K}_* = \mathbf{K}(\boldsymbol{t}_*, \boldsymbol{t}) \in \mathbb{R}^{N_p \times N_t}\), and \(\mathbf{K}_{**} = \mathbf{K}(\boldsymbol{t}_*, \boldsymbol{t}_*)\). 
Conditioned on observations \(\boldsymbol{y}\), the posterior distribution over the other points is given by:
\begin{equation}
\boldsymbol{f}_* \mid \boldsymbol{t}, \boldsymbol{y}, \boldsymbol{t}_* \sim \mathcal{N}\left( \hat{\boldsymbol{f}}(\boldsymbol{t}_*), \operatorname{Cov}(\boldsymbol{f}_*) \right),
\end{equation}
with posterior mean and covariance:
\begin{align}
\hat{\boldsymbol{f}}(\boldsymbol{t}_*) &= \mathbf{K}_* (\mathbf{K} + \boldsymbol{\Sigma})^{-1} \boldsymbol{y}, \\
\operatorname{Cov}(\boldsymbol{f}_*) &= \mathbf{K}_{**} - \mathbf{K}_* (\mathbf{K} + \boldsymbol{\Sigma})^{-1} \mathbf{K}_*^\top.
\end{align}

Leaving observations with zero mean \(\boldsymbol{y} = \boldsymbol{0}\) at specificway points, which results in \(\hat{\boldsymbol{f}}(\boldsymbol{t}_*) = \boldsymbol{0}\). 
The diagonal entries of \(\operatorname{Cov}(\boldsymbol{f}_*)\) represent the posterior variances at the test points can be obtained, denoted as \(\boldsymbol{\sigma}^2(\boldsymbol{t}_*)\), and are directly used as interpolated tolerance bounds for the tracking trajectory.

\appendix
\section*{Appendix B}
\subsection*{Solving Original MINLP with the Novel Mapping Function}
\label{cmaes_anylsis_MINLP}

The fundamental principle of the CMA-ES algorithm~\citep{hansen2003reducing} is to iteratively sample candidate solutions from a multivariate normal distribution, whose mean and covariance are adaptively updated to guide the search for the local optimum. At each iteration, this distribution is centred around the current estimate of the optimum and updated based on the fitness values of the samples. Consider the task of optimizing the placement of two modules, labelled 1 and 2. The optimization variables include two continuous parameters, \(x\) and \(y\), which correspond to the modules, along with an additional continuous parameter \(z\). Thus, CMA-ES operates in a three-dimensional space defined by \((x, y, z)\)(see Fig.~\ref{sample_example}, left), sampling candidates iteratively.

When projecting the sampling process onto the \(x\)-\(y\) plane (see Fig.~\ref{sample_example}, right), the algorithm explores possible combinations of \(x\) and \(y\) and their associated scores. With a designed mapping, each continuous sample \(\bm{M} = [x,\, y]^\top\) is mapped to a discrete morphology as follows:
\[
g(\bm M) =
\begin{cases}
[2,1], & y > x \\
[1,2], & y < x
\end{cases}
\]
(as illustrated in Fig.~\ref{two_variable_situation_case}). Here, the arrangement \([1, 2]\) is chosen when \(x > y\) (points below the decision boundary), while \([2, 1]\) is chosen for \(x < y\) (points above the line). Through this mapping, the continuous sampling of CMA-ES effectively solves a discrete arrangement problem by converging the distribution toward the region encoding the optimal morphology.

\begin{figure}[h]
  \centering
\includegraphics[width=0.45\textwidth]{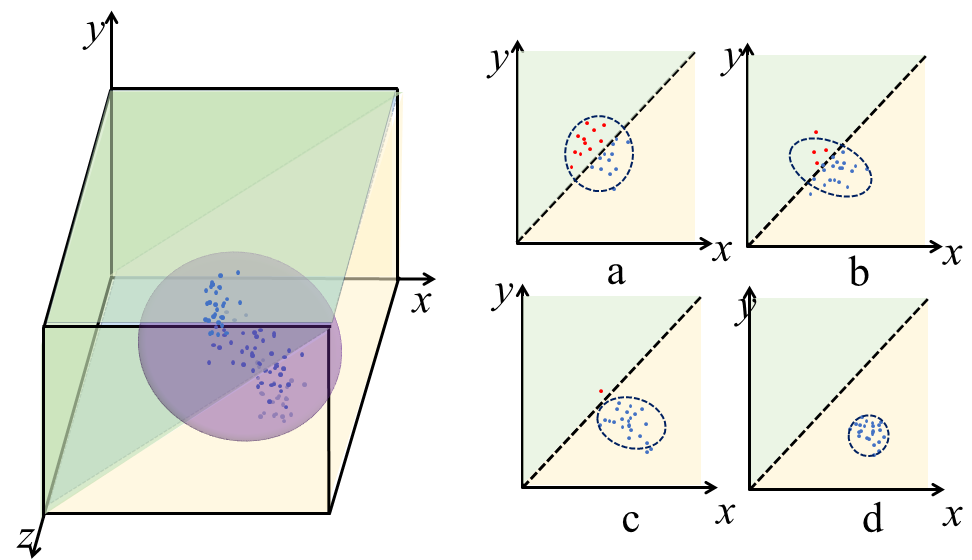}
\caption{\textbf{Sampling and evaluation process in 3D and 2D.} (Left) Sampling candidate solutions in the 3D space at one step. (Right) The evaluation process is projected onto the 2D \(x\)-\(y\) plane for analysing candidate solutions.}
  \label{sample_example}
\end{figure}

\begin{figure}[h]
  \centering
  \includegraphics[width=0.35\textwidth]{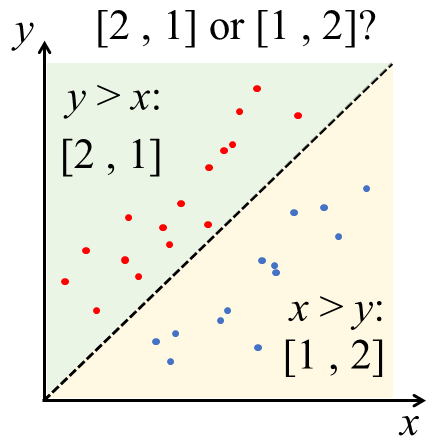}
\caption{\textbf{Mapping continuous states \((x, y)\) to discrete morphology states.} The region \(y > x\) (green) yields the arrangement \([2, 1]\), while \(y < x\) (yellow) yields \([1, 2]\). Red dots indicate sampled states, and the dashed line denotes the decision boundary.}

  \label{two_variable_situation_case}
\end{figure}

\end{document}